\documentclass[10pt,twocolumn,letterpaper]{article}

\usepackage{cvpr}              % To produce the CAMERA-READY version
\usepackage[accsupp]{axessibility}
\usepackage{multirow}
\usepackage[dvipsnames,table]{xcolor}
\usepackage{dsfont} 
\newcommand{\red}[1]{{\color{red}#1}}

\newcommand{\supp}{{supplementary material}\ }
\newcommand{\suppeos}{{supplementary material}}

\newcommand{\ours}{{\text{KEPP}}\ }
\newcommand{\ourseos}{{\text{KEPP}}} 

\definecolor{cvprblue}{rgb}{0.21,0.49,0.74}
\usepackage{color,soul}
\usepackage[pagebackref,breaklinks,colorlinks,citecolor=cvprblue]{hyperref}
\usepackage{booktabs} % For prettier tables
\usepackage{amssymb} % For less than equal to symbol
\usepackage{graphicx} % Required for including images

\DeclareMathAlphabet{\mathmybb}{U}{bbold}{m}{n}

\def\paperID{12890} % *** Enter the Paper ID here
\def\confName{CVPR}
\def\confYear{2024}

\title{Why Not Use Your Textbook? Knowledge-Enhanced Procedure Planning of Instructional Videos}

\author{Kumaranage Ravindu Yasas Nagasinghe\textsuperscript{1}
\and
Honglu Zhou\textsuperscript{2}
\and
Malitha Gunawardhana\textsuperscript{1,3} 
\and 
Martin Renqiang Min\textsuperscript{2} 
\and
Daniel Harari\textsuperscript{4}
\and
Muhammad Haris Khan\textsuperscript{1}
\and
\small{
\textsuperscript{1}Mohamed bin Zayed University of Artificial Intelligence,   \textsuperscript{2}NEC Laboratories, USA, } \\
\small{\textsuperscript{3} University of Auckland, \textsuperscript{4}Weizmann Institute of Science}\\
{\tt\small \href{ravindu.nagasinghe@mbzuai.ac.ae}{ravindu.nagasinghe@mbzuai.ac.ae}, 
\href{muhammad.haris@mbzuai.ac.ae}{muhammad.haris@mbzuai.ac.ae} }
}

\begin{document}
\maketitle
\begin{abstract}

In this paper, we explore the capability of an agent to construct a logical sequence of action steps, thereby assembling a strategic procedural plan. This plan is crucial for navigating from an initial visual observation to %the
a target visual outcome, as depicted in real-life instructional videos. Existing works have attained partial success by extensively leveraging various sources of information available in the datasets, such as heavy intermediate visual observations, procedural names, or natural language step-by-step instructions, for features or supervision signals. However, the task remains formidable due to the implicit causal constraints in the sequencing of steps and the variability inherent in multiple feasible plans. To tackle these intricacies that previous efforts have overlooked, we propose to enhance the agent's capabilities by infusing it with procedural knowledge. This knowledge, sourced from %the 
training procedure plans and structured as a directed weighted graph, equips the agent to better navigate the complexities of step sequencing and its potential variations. We coin our approach %as 
\ourseos, a novel Knowledge-Enhanced Procedure Planning system, which harnesses a probabilistic procedural knowledge graph extracted from %the 
training data, effectively acting as a comprehensive textbook for the training domain. Experimental evaluations across three widely-used datasets under settings of varying complexity reveal that \ours attains superior, state-of-the-art results while requiring only minimal supervision. Code and trained model are available at {{\fontsize{8}{12} \tt  \href{https://ravindu-yasas-nagasinghe.github.io/KEPP-Project_Page/}{https://github.com/Ravindu-Yasas-Nagasinghe/KEPP} }}

\end{abstract}
\vspace{-4mm}
\section{Introduction}
\label{sec:intro}

\begin{figure}
    \centering
    \includegraphics[width=\linewidth]{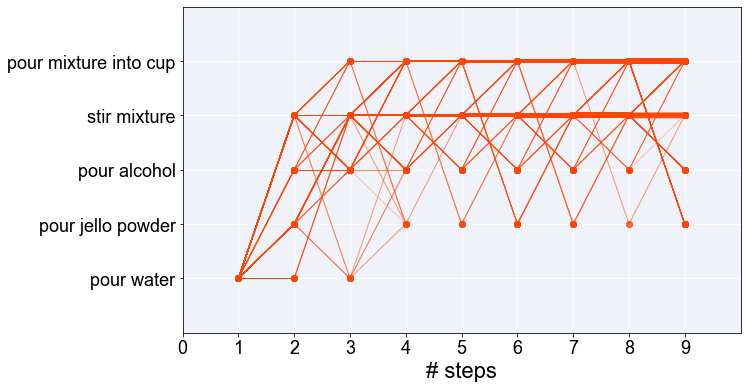}
    \vspace{-10pt}
    \caption{\textbf{Expert trajectories~\cite{bi2021procedure} of the `Make Jello Shots' task from the CrossTask dataset~\cite{zhukov2019cross}.} Heavier color indicates more frequently visited path. This depicts the complexities of the procedure planning task, arising from the subtle causal links in step sequencing (e.g., steps like `stir mixture' or `pour mixture' typically occur after adding individual ingredients), the varied probabilities of transitioning between steps, and the diversity in plans viable for a given starting point and an intended outcome. Motivated by these nuanced challenges, we propose Knowledge-Enhanced Procedure Planning (\ourseos) with the use of a probabilistic procedure knowledge graph to capture and represent these intricacies
    }
    \label{fig:expert_tra_jelo}
\end{figure}

The evolution of the internet has precipitated an unprecedented influx of video content, serving as a vital educational resource for myriad learners. Individuals frequently leverage platforms such as YouTube to acquire new skills, ranging from culinary arts to automobile maintenance~\cite{miech2019howto100m}. While these instructional videos benefit the development of intelligent agents in mastering long-horizon tasks, the challenge 
% for embodied agents like robots
extends beyond merely interpreting visuals. It requires high-level reasoning and planning to effectively assist in complex, real-life scenarios~\cite{du2023video}.

Procedure Planning in Instructional Videos
% is a fundamental task that
demands an agent to produce a sequence of actionable steps, thereby crafting a procedure plan that facilitates the transition from %the 
an initial visual observation of the physical world towards achieving %the
a desired goal state~\cite{chang2020procedure,bi2021procedure,sun2022plate,zhao2022p3iv,wang2023pdpp,wang2023event}.
% ,schema}.
The task acts as a precursor to an envisioned future scenario in which an agent like a robot provides on-the-spot support, such as assisting an individual in preparing a recipe~\cite{activepp}. 
% For embodied agents like robots, the ability, gained through instructional video-based procedure planning, to perceive the state of the world and devise coherent action plans in conjunction with or on behalf of humans appears indispensable~\cite{activepp}.
% For embodied agents like robots, indispensably require the skill to comprehend the world and formulate cohesive action plans with or for humans, a capability that can be developed through instructional video-based procedure planning~\cite{activepp}.

Current methods in procedure planning in instructional videos make extensive use of various annotations available within the datasets to enrich input features or provide supervisory signals (see Table~\ref{tab:cross-task}). These include detailed, temporally localized visual observations of intermediate action steps throughout the procedure plan~\cite{chang2020procedure,bi2021procedure,sun2022plate}, high-level procedural task labels~\cite{wang2023pdpp,wang2023event},
% ,schema}, 
and step-by-step instructions in natural language~\cite{zhao2022p3iv,wang2023event}.
% ,schema}.
Despite advancements, significant challenges persist, including characterizing the implicit causal constraints in step sequencing, the varied probabilities of transitioning between steps, and the inherent variability of multiple viable plans (see Fig.~\ref{fig:expert_tra_jelo}).

To address these intricacies that previous efforts have overlooked, 
% In response,
we propose to enhance the agent's capabilities in procedure planning by infusing it with comprehensive procedural knowledge~\cite{zhou2023procedure}, derived from %the 
training procedure plans and structured %into 
as a directed weighted graph. 
This graph, as a Probabilistic Procedural Knowledge Graph~\cite{ashutosh2023video} where nodes denote steps from diverse tasks and edges represent step transition probabilities in the training domain, 
empowers agents to more adeptly navigate the complexities of step sequencing and its potential variations. 

\begin{figure*}[t]
	\centering
	\vspace{-10pt}
    \includegraphics[width=\textwidth]{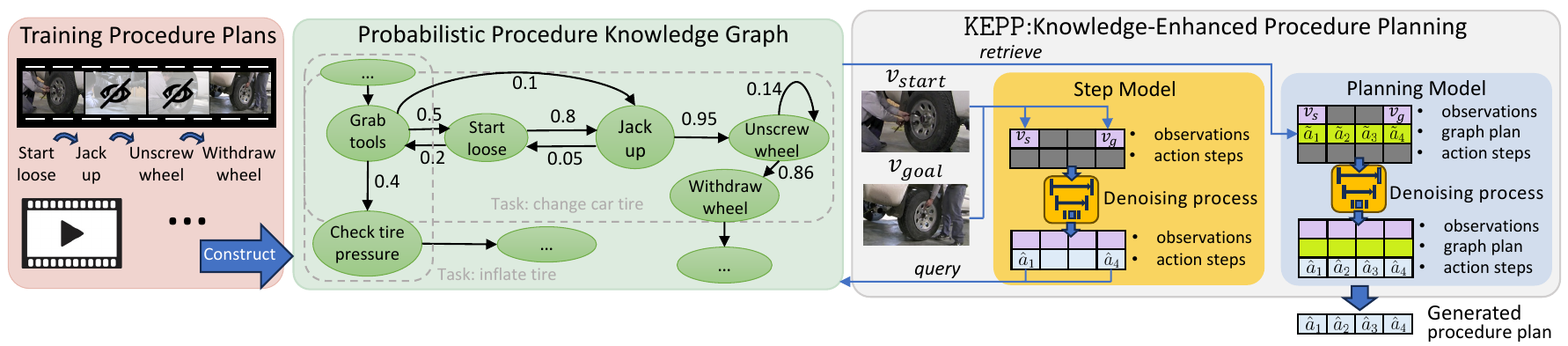}
    \vspace{-15pt}
	\caption{
	\textbf{Overview of our methodology.}  We introduce \ourseos, a Knowledge-Enhanced Procedure Planning system for instructional videos, leveraging a Probabilistic Procedural Knowledge Graph (P$^2$KG). \ours 
 % requires minimal annotations
  breaks down procedure planning into two parts: predicting initial and final steps from visual states, and crafting a procedure plan based on the procedural knowledge retrieved from P$^2$KG, conditioned on the predicted first and last action steps. 
 \ours requires minimal annotations and enhances planning effectiveness
}
\label{fig:model_overview}
\end{figure*}

Our proposed approach, \ourseos, is a novel Knowledge-Enhanced Procedure Planning system (Fig.~\ref{fig:model_overview}) that harnesses a probabilistic procedural knowledge graph (P$^2$KG), constructed from training procedure plans. This graph functions like a detailed textbook, providing extensive knowledge for the training domain, and thereby circumventing the need for costly multiple annotations required by existing methods. Additionally, we decompose the instructional video procedure planning problem into two parts: one driven by objectives specific to step perception and the other by a procedural knowledge-informed modeling of procedure planning. In this problem decomposition, the first and last action steps are predicted based on the initial and goal visual states. Following this, a procedure plan is crafted by leveraging the procedure plan recommendations retrieved from the P$^2$KG. The recommendations correspond to the most probable procedure plans frequently used in training, conditioned on the predicted first and last action steps. In a similar vein to the approach by Li~\etal~\cite{li2023skip}, our proposed decomposition strategy reduces uncertainty by maximizing the use of currently available information, namely the initial and goal visual states. This allows for the improvement of procedure planning through more accurate predictions of start and end actions. Plus, this decomposition effectively incorporates 
procedural knowledge 
into procedure planning, thereby enhancing its effectiveness. 

Our contributions are as follows:
\begin{itemize}
    \item  We propose \ourseos, a Knowledge-Enhanced Procedure Planning system for instructional videos that leverages rich procedural knowledge from a probabilistic procedural knowledge graph (P$^2$KG). This approach necessitates only a minimal amount of annotations for supervision.
    \item We decompose the problem in procedure planning of instructional videos: predicting the initial and final steps from the start and end visuals, and then creating a plan using procedural knowledge retrieved based on these predicted steps. This approach prioritize the currently available information and effectively incorporates procedural knowledge, enhancing strategic planning.
    \item Experimental evaluations on three widely-used datasets, under settings of varying complexity, reveal that \ours attains state-of-the-art results in procedure planning. 
    %Code and trained models will be made publicly available.
\end{itemize}

\section{Related Work}

\noindent \textbf{Instructional Videos},
which demonstrate multi-step procedures, have become a hotbed of research. The studies delve into various aspects, including comprehending and extracting intricate spatiotemporal content from video~\cite{flanagan2023learning,ghoddoosian2023weakly,hu2023compositional,fried2020learning,souvcek2022look,moltisanti2023learning,shah2023steps,zala2023hierarchical,doughty2020action,shen2021learning,xu2020benchmark}, interpreting the interrelationships between various actions and procedural events~\cite{song2023ego4d,zhukov2019cross}, 
and developing capabilities for forecasting~\cite{sener2022transferring,nawhal2022rethinking} and strategic
reasoning and 
planning~\cite{huang2022inner} within the context of these videos. Furthermore, by leveraging the multimodality of visual, auditory, and narrative elements within these videos, research extends to areas like multimodal alignment~\cite{afouras2023ht,zhang2023aligning}, grounding~\cite{chen2023and,mavroudi2023learning,dvornik2022flow,tan2021look,huang2018finding}, representation learning~\cite{miech2020end,dong2023weakly,zhong2023learning}, pre-training~\cite{huang2020multimodal,zhou2023procedure,dvornik2023stepformer}, and more~\cite{narasimhan2022tl,fischer2022vilt,yang2021induce,huang2017unsupervised}. This paper focuses on procedure planning in instructional videos.

\noindent \textbf{Procedure Planning} 
is a vital skill for autonomous agents tasked with handling complex activities in everyday settings. Essentially, these agents must discern the appropriate actions to reach a specific goal. This aspect of artificial intelligence (AI) has been a prominent and integral subject in fields like robotics~\cite{singh2023progprompt,shridhar2020alfred,huang2022inner,gan2021threedworld, mao2023action}. Yet, the challenge of procedure planning in the context of instructional videos is notably distinct, and potentially more complex, than its counterparts in natural language processing~\cite{lu2022neuro,brahman2023plasma}, multimodal generative AI~\cite{lu2023multimodal,du2023video}, and simulated environments~\cite{shridhar2020alfred,huang2022inner,huang2022language}.
% , and human pedestrian or vehicle trajectory planning~\cite{}. 
Its significance is underscored by the need for planning that is grounded in real-world scenes. This requires the development of AI agents capable of accurately perceiving and understanding the current real-world context, and then anticipating and mapping out a logical sequence of actions to fulfill a high-level goal effectively.

\noindent \textbf{Procedure Planning in Instructional Videos} has recently garnered research attention. DDN~\cite{chang2020procedure} initiates this trend by conceptualizing the problem as sequential latent space planning. Building on this, PlaTe~\cite{sun2022plate} employs transformers for both action and state models, integrating Beam Search for enhanced performance. Meanwhile, Ext-GAIL~\cite{bi2021procedure} suggests employing contextual modeling through variational autoencoder and adversarial policy learning. This method considers contextual information as time-invariant knowledge, crucial for distinguishing specific tasks and allowing for multiple planning outcomes.

While these earlier approaches have viewed procedure planning as an autoregressive sequence generation problem, recent methods regard it as a distribution-fitting problem to mitigate error propagation in sequential decisions. In this vein, P$^3$IV~\cite{zhao2022p3iv} replaces intermediate visual states with linguistic representations for supervision, predicting all steps simultaneously instead of using autoregressive methods. To circumvent the complex learning strategies and high annotation costs of previous work, PDPP~\cite{wang2023pdpp} models the entire intermediate action sequence distribution using a conditioned projected diffusion model. This approach redefines the planning problem as a sampling process from this distribution and simplifies supervision by using only instructional video task labels. E3P~\cite{wang2023event}, also encoding task information, adopts a mask-and-predict strategy for mining step relationships
in procedural tasks, 
integrating probabilistic masking for regularization. 
In contrast, our approach does not rely on annotations of intermediate states, natural language step representations, or procedural task labels.

% Acknowledging
Recognizing the difficulties inherent in high dimensional state supervision and the accumulation of errors in action sequences, SkipPlan~\cite{li2023skip} was developed. It strategically focuses on action predictions, breaking down longer sequences into shorter, more manageable sub-chains by skipping over less reliable intermediate actions. Drawing inspiration from SkipPlan, our approach decomposes the procedure planning problem to prioritize the most reliable information available (\textit{ref.} \S~\ref{subsubsec:problem_decomposition}). However, we innovate further by incorporating a Probabilistic Procedure Knowledge Graph, significantly enriching the planning phase.
% \clearpage

% \vspace{-5pt}
\section{Methodology}
We will first introduce the problem setup in \S~\ref{subsec:problem_formulation}, and then discuss our novel Knowledge-Enhanced Procedure Planning system (\ourseos) in \S~\ref{subsec:kepp}. See Fig.~\ref{fig:model_overview} for \ours overview.

\subsection{Problem and Method Overview}
\label{subsec:problem_formulation}

% \noindent \textbf{Problem Formulation.}
\subsubsection{Problem Formulation}
\label{subsubsec:problem_formulation}
We follow the problem definition for procedure planning of instructional videos put forth by Chang \etal~\cite{chang2020procedure}: given an observation of the initial state  $v_{start}$ and a goal state $v_{goal}$, both are short video clips indicating different states of the real-world environment extracted from an instructional video, a model is required to plan a sequence of action steps $a_{1:T}$ 
% with a predefined planning horizon $T$ so that the environment state can be transformed from $v_{start}$ to $v_{goal}$.
to reach the indicated goal.
Here, $T$ is the planning horizon, inputting to the model, corresponding to the number of action steps in the sequence produced by the model so that the environment state can be transformed from $v_{start}$ to $v_{goal}$. We use $a_t$ to denote the action step at the timestamp $t$, and in the following, $v_{s}$ and $v_{g}$ are short for $v_{start}$ and $v_{goal}$. 
Mathematically, the procedure planning problem is defined as $p\left(a_{1:T}|v_s, v_g\right)$ that denotes the conditional probability distribution of the action sequence $a_{1:T}$ given the initial visual observation $v_{start}$ and the goal visual state $v_{goal}$.

\vspace{-10pt}
\subsubsection{Problem Decomposition}
\label{subsubsec:problem_decomposition}
Considering
the initial and final visual states are
input, 
providing the most reliable information,
we hypothesize that predicting the first and final action steps is more dependable than 
interpolating
the intermediate ones, and consequently, an enhanced accuracy in predicting the first and final steps can lead to more effective procedure planning. Inspired by this hypothesis, we decompose the procedure planning problem into two sub-problems, as shown in Eq.~\ref{eq:problem_decompose}:
\begin{equation}
% \small
    p\left(\hat{a}_{1: T}|v_s, v_g\right) = p\left(\hat{a}_{2:T-1}|\hat{a}_1, \hat{a}_T\right) p\left(\hat{a}_1, \hat{a}_T|v_s, v_g\right),
    \label{eq:problem_decompose}
\end{equation}
where  
the first sub-problem is to identify the beginning step $a_1$ and the end step $a_T$, and the second sub-problem is to plan the intermediate action steps $a_{2:T-1}$ given $a_1$ and $a_T$. We use $\hat{a}_t$ to denote \textit{predicted} action step at timestamp $t$.

Our proposed problem decomposition
% formulation 
in Eq.~\ref{eq:problem_decompose}
bears resemblance with the problem formulation from Li~\etal~\cite{li2023skip};
they decompose procedure planning into $p\left(\hat{a}_{1: T}|v_s, v_g\right)=\prod_{t=2}^{T-1} p\left(\hat{a}_t|\hat{a}_1, \hat{a}_T\right) p\left(\hat{a}_1, \hat{a}_T|v_s, v_g\right)$. 
However, our formulation differs in its approach to modeling the second sub-problem. Specifically, we employ a conditioned projected diffusion model (\textit{ref.}~\S~\ref{subsec:kepp}) to jointly predict $a_{2:T-1}$ at once, whereas Li~\etal~\cite{li2023skip} rely on Transformer decoders to predict each intermediate action independently. Further, we integrate a Probabilistic Procedure Knowledge Graph (\textit{ref.}~\S~\ref{subsubsec:p2kg}) to address the second sub-problem.

Tackling the second sub-problem is nontrivial even when armed with an oracle predictor for the first sub-problem. Procedure planning in real-life scenarios remains daunting because of the following \textbf{challenges}: \textit{(1)} the presence of implicit temporal and causal constraints in the sequencing of steps, \textit{(2)} the existence of numerous viable plans given an initial state and a goal state, and \textit{(3)} the need to incorporate the real-life everyday knowledge both in task-sharing steps and in managing the inherent variability in transition probabilities between steps.
Previous studies tackled these challenges by extensively harnessing detailed annotations in the datasets to augment input features or offer supervision signals (see Table~\ref{tab:cross-task}).
In contrast, we propose harnessing a Probabilistic Procedural Knowledge Graph (P$^2$KG) which is extracted from the procedure plans in the training set. 
With the P$^2$KG at our hand, 
we further decompose the procedure planning problem 
to reduce its complexity and learn
$f_\theta:(v_s, v_g, T) \rightarrow p\left(\hat{a}_{1: T}|v_s, v_g\right)$
as follows:

\begin{equation}
% \small
\begin{array}{l}
    p\left(\hat{a}_{1: T}|v_s, v_g\right) = \\
    p\left(\hat{a}_{1:T}|\tilde{a}_{1:T}, v_s, v_g\right)  p\left(\tilde{a}_{1:T}|\hat{a}_1, \hat{a}_T\right) p\left(\hat{a}_1, \hat{a}_T|v_s, v_g\right)
    \label{eq:problem_decompose_p2kg}
\end{array}
\end{equation}
where $f_\theta$ denotes the machine learning model, and $\tilde{a}_{1:T}$ represents a graph path (i.e., a sequence of nodes) retrieved from P$^2$KG. This graph path provides a valuable procedure plan recommendation aligned with the training domain, thus mitigating the complexity of procedure planning.
It is worth noting that the proposed approach to modeling procedure planning using Eq.~\ref{eq:problem_decompose_p2kg} demands only a minimal level of supervision, 
merely relying on the ground truth training procedure plan;
Eq.~\ref{eq:problem_decompose_p2kg} circumvents the need for additional annotations.
We describe details of our P$^2$KG-enhanced approach in the following subsection.

\subsection{\ourseos: Knowledge-Enhanced Procedure Planning}
\label{subsec:kepp}

We propose \ours
(Fig.~\ref{fig:model_overview}) utilizing a probabilistic procedure knowledge graph extracted from the training set. We firstly identify the beginning and conclusion steps according to the input initial and goal states; and then, conditioned on these steps and the planning horizon $T$, we query the graph to retrieve relevant procedural knowledge for knowledge-enhanced procedure planning of instructional videos. 
 
\subsubsection{Identify Beginning and Conclusion Steps}
Given $v_{start}$ and $v_{goal}$ as input, we adapt a Conditioned Projected Diffusion Model~\cite{wang2023pdpp} (\textit{ref.}~\suppeos) to identify the first action step and the final step; we refer to this model as the `Step (Perception) Model'.

\noindent \textbf{Standard Denoising Diffusion Probabilistic Model}
tackles data generation through a denoising Markov chain over variables $\left\{x_N \ldots x_0\right\}$, starting with $x_N$ as a Gaussian random distribution~\cite{ho2020denoising}.
In the forward diffusion phase, Gaussian noise $\epsilon \sim \mathcal{N}(0, \mathbf{I})$ is progressively added to the initial, unaltered data $x_0$, transforming it into a Gaussian random distribution. 
Conversely, the reverse denoising process transforms Gaussian noise back into a sample. Denoising is parameterized by a learnable noise prediction model, and the learning objective is to learn the noise added to $x_0$ at each diffusion step.
After training, the diffusion model generates data akin to $x_0$ by iteratively applying the denoising process, starting from random Gaussian noise.

\noindent \textbf{Adopting Conditioned Projected Diffusion Model as the Step Model.} 
For our step model, the distribution we aim to fit is the two-action sequence $\left[a_1, a_T\right]$, based on the
visual
initial and goal states, $v_{start}$ and $v_{goal}$. These conditional visual states are concatenated with the actions along the action feature dimension, forming a multi-dimensional array:
% \vspace{-20pt}
\begin{equation}
    \left[\begin{array}{ccccc}
\cellcolor{gray!25}v_s & \cellcolor{gray!25}0 & \cellcolor{gray!25}\ldots & \cellcolor{gray!25}0 & \cellcolor{gray!25}v_g\\
a_1 & \cellcolor{gray!25}0 & \cellcolor{gray!25}\ldots & \cellcolor{gray!25}0 & a_T 
\end{array}\right]
% \vspace{-20pt}
\end{equation}
where the
% number of columns in this array corresponds to the planning horizon $T$. The
array is zero-padded to have a length corresponds to the planning horizon $T$.
During the denoising process, these conditional visual states can change, potentially misleading the learning process. To prevent this, a condition projection operation~\cite{wang2023pdpp} is applied, ensuring the visual state and zero-padding dimensions remain unchanged (shaded below).
The projection operation is denoted as:
% \vspace{-10pt}
\begin{equation}
\footnotesize
  \left[\begin{array}{ccccc}
\cellcolor{gray!25}\hat{v}_1 & \cellcolor{gray!25}\hat{v}_2 & \cellcolor{gray!25}\ldots & \cellcolor{gray!25}\hat{v}_{T-1} & \cellcolor{gray!25}\hat{v}_T \\
\hat{a}_1 & \cellcolor{gray!25}\hat{a}_2 & \cellcolor{gray!25}\ldots & \cellcolor{gray!25}\hat{a}_{T-1} & \hat{a}_T 
\end{array}\right] \xrightarrow[\text{ }]{\text{Projection}} \left[\begin{array}{ccccc}
\cellcolor{gray!25}v_s & \cellcolor{gray!25}0 & \cellcolor{gray!25}\ldots & \cellcolor{gray!25}0 & \cellcolor{gray!25}v_g \\
\hat{a}_1 & \cellcolor{gray!25}0 & \cellcolor{gray!25}\ldots & \cellcolor{gray!25}0 & \hat{a}_T 
\end{array}\right]
\end{equation}
where $\hat{v}_t$ denotes the predicted visual state dimensions at timestamp $t$ within the planning horizon $T$.

\subsubsection{Construct the Probabilistic Procedure Knowledge Graph (P$^2$KG)}
\label{subsubsec:p2kg}
% \subsubsection{Constructing the P$^2$KG}
The Probabilistic Procedure Knowledge Graph~\cite{ashutosh2023video} $\text{P$^2$KG}=(V, E, w)$ is 
% constructed from a collection of training procedure plans, forming 
a directed and weighted graph. In this structure, each step 
% (which can belong to multiple tasks) 
from the training set is represented as a node. During the graph construction process, we iterate over the training procedure plans, and for each direct step transition present in a plan, we add an edge from $a_t$ to $a_{t+1}$ if it does not already exist in the graph; otherwise, we increase its existing frequency count by one. Eventually, this process results in a frequency-based Procedural Knowledge Graph (PKG)~\cite{zhou2023procedure}, which adeptly encapsulating the complexities of step sequencing in procedures and its potential variations, thereby addressing challenges \textit{(1)} and \textit{(2)} of procedure planning (\textit{ref.}~\S~\ref{subsubsec:problem_decomposition}).
To further tackle challenge \textit{(3)}, this graph undergoes a transformation into a probabilistic format. In this transformed graph, the edges are not just connections but also signify the likelihoods of transitioning from one step to another. The weight of an edge from $a_t$ to $a_{t+1}$ is the count of transitions from action step $a_t$ to $a_{t+1}$ normalized by total count of $a_t$ being executed~\cite{ashutosh2023video}. The normalization converts the frequency-based weight into probability distribution and the sum of all out-going edges is one.

\subsubsection{P$^2$KG-Enhanced Procedure Planning}
\label{subsubsec:planning-model-ref}
\noindent \textbf{Retrieving Procedure Plan Recommendations from the  P$^2$KG.} 
Humans use both previously-acquired knowledge and external knowledge when solving problems. The P$^2$KG provides extensive procedural knowledge, serving as a comprehensive textbook, particularly beneficial for the planning model that requires advanced skills. 

To utilize this procedural knowledge, queries are made to the P$^2$KG using the first ($\hat{a}_1$) and last ($\hat{a}_T$) actions predicted by the step model. The aim is to find graph paths no longer than $T$ steps, starting from $\hat{a}_1$ and ending at $\hat{a}_T$. 
This above process often results in multiple possible paths. To evaluate these paths, the probability of each is calculated by multiplying the probability weights of the edges along the path. For instance, the probability of a path $a_1 \rightarrow a_2 \rightarrow a_3$ is determined by the product $w_{a_1 \rightarrow a_2} \times w_{a_2 \rightarrow a_3}$. These paths are then ranked according to their probabilities, and the top $R$ paths are selected as the recommended procedure plans from the P$^2$KG, where $R$ is predefined.
For paths shorter than $T$, padding is applied at any point in the middle of the sequence to explore all possible resultant paths. When $R$ is greater than one, the top $R$ paths are aggregated through linear weighting into a single path (See section A.2 of supplementary material). This final path is then used as an additional input for the procedure planning model, thereby enhancing its decision-making process.

\noindent \textbf{Adopting Conditioned Projected Diffusion Model as the Planning Model.} 
For the planning model, the conditional visual states and the procedure plan recommendation from the P$^2$KG are concatenated with the actions along the action feature dimension, forming a multi-dimensional array:
% \vspace{-20pt}
\begin{equation} \label{eq:matrix-multi}
    \left[\begin{array}{ccccc}
\cellcolor{gray!25}v_s & \cellcolor{gray!25}0 & \cellcolor{gray!25}\ldots & \cellcolor{gray!25}0 & \cellcolor{gray!25}v_g\\
\cellcolor{gray!25}\tilde{a}_1 & \cellcolor{gray!25}\tilde{a}_2 & \cellcolor{gray!25}\ldots & \cellcolor{gray!25}\tilde{a}_{T-1} & \cellcolor{gray!25}\tilde{a}_T\\
a_1 & a_2 & \ldots & a_{T-1} & a_T 
\end{array}\right]
% \vspace{-20pt}
\end{equation}

The rest process is similar to the step model, except that the project operation guarantees that three specific aspects remain unaltered--the dimensions of the the visual state, P$^2$KG recommendation, and zero-padding.
\section{Experiments}

\begin{table*}[!htp]
\setlength{\tabcolsep}{2.6pt}
\small
  \aboverulesep=0ex
  \belowrulesep=0ex 
\centering
\begin{tabular}{l|cccc|ccc|ccc}
\toprule
\multirow{2}{*}{\textbf{Models}}                              &\multicolumn{4}{c|}{\textbf{Required Annotations}}&\multicolumn{3}{c|}{\( T=3 \)} & \multicolumn{3}{c}{\( T=4 \)} \\ \cline{2-11} 
                                    &step class&  visual states& step text& task class&\( SR^{\uparrow} \) & \( mAcc^{\uparrow} \) & \( mIoU^{\uparrow} \) & \( SR^{\uparrow} \) & \( mAcc^{\uparrow} \) & \(mIoU^{\uparrow} \) \\ \midrule
Random                              &\checkmark&  & & 

&\( <0.01 \) & 0.94  & 1.66  & \( <0.01 \)   & 0.83     & 1.66           \\
Retrieval-Based                     &\checkmark&  & & 

&8.05& 23.3& 32.06& 3.95& 22.22& 36.97
\\
WLTDO   \cite{ehsani2018let}          &\checkmark&  \checkmark& & 

&1.87& 21.64& 31.70& 0.77& 17.92& 26.43
\\
UAAA   \cite{abu2019uncertainty}      &\checkmark&  \checkmark& & 

&2.15& 20.21& 30.87& 0.98& 19.86& 27.09
\\
UPN   \cite{srinivas2018universal}    &\checkmark&  \checkmark& & 

&2.89& 24.39& 31.56& 1.19& 21.59& 27.85
\\
DDN   \cite{chang2020procedure}       &\checkmark&  \checkmark& & 

&12.18& 31.29& 47.48& 5.97& 27.10& 48.46
\\
PlaTe   \cite{sun2022plate}        &\checkmark&        \checkmark& & 

&16.00&      36.17&      65.91&     14.00&     35.29&     55.36
\\
Ext-GAIL wo Aug. \cite{bi2021procedure}     &\checkmark&  \checkmark& & 

&18.01& 43.86& 57.16& -& -& -
\\
Ext-GAIL \cite{bi2021procedure}             &\checkmark&  \checkmark& & 

&21.27& 49.46& 61.70& 16.41& 43.05& 60.93
\\
P$^3$IV $^\clubsuit$  \cite{zhao2022p3iv}                &\checkmark&  &\checkmark& 

&23.34& 49.96& 73.89& 13.40& 44.16& 70.01
\\

PDPP $^\clubsuit$   \cite{wang2023pdpp}     &\checkmark&  & & \checkmark
&26.38& 55.62& 59.34& 18.69& 52.44& 62.38
\\ 
E3P $^\clubsuit$   \cite{wang2023event}        &\checkmark&        & \checkmark& 
\checkmark
&26.40&      53.02&      74.05&     16.49&     48.00&     70.16
\\
SkipPlan  \cite{li2023skip} $^\clubsuit$       &\checkmark&        & & 

&28.85&      61.18&      \textbf{74.98} &     15.56&     55.64&     \textbf{70.30}
\\ \midrule \midrule
Ours    w/ P$^2$KG ($R$=2)     &\checkmark&        & & 

&22.60&      48.76&      53.57&     13.90&   45.79  &     55.00
\\ 
Ours $^\clubsuit$  w/ P$^2$KG ($R$=1) &\checkmark&        & & 

&33.34&      \textbf{61.36}&      64.14&     20.38&     55.54&     64.03\\
Ours $^\clubsuit$   w/ P$^2$KG ($R$=2)     &\checkmark&        & & 

&\textbf{33.38}&      60.79 &      63.89&     \textbf{21.02} &     \textbf{56.08} &     64.15\\ \midrule

PDPP $^\clubsuit$  {\textdagger}  \cite{wang2023pdpp}   &\checkmark&    & & \checkmark
&37.20 &  64.67&   66.57&   21.48&     57.82&       65.13
\\ 
Ours $^\clubsuit$  {\textdagger}    w/ P$^2$KG ($R$=1)            &\checkmark&   & & &\textbf{38.12}&   \textbf{64.74}&   \textbf{67.15}&   \textbf{24.15}&   \textbf{59.05}&     \textbf{66.64}
\\

\bottomrule

\end{tabular}
\caption{Performance of our method in comparison to existing baselines for CrossTask dataset. $^\clubsuit$ means that the input visual features are from the S3D network~\cite{miech2020end} pretrained on HowTo100M~\cite{miech2019howto100m}; otherwise, precomputed features provided in CrossTask are used. {\textdagger} indicates the results are under the PDPP's task setting, while others are under the conventional setting 
}
\label{tab:cross-task}
\end{table*}

\begin{table}[!htp]
\setlength{\tabcolsep}{3.6pt}
\small
  \aboverulesep=0ex
  \belowrulesep=0ex 
\centering
\vspace{-10pt}
\begin{tabular}{lcc}
\toprule
Models                                & \( T=5 \) & \( T=6 \)\\ 
\midrule
                                   
DDN  \cite{chang2020procedure}       & 3.10 & 1.20\\
P$^3$IV $^\clubsuit$  \cite{zhao2022p3iv}           & 7.21 & 4.40\\
PDPP $^\clubsuit$  \cite{wang2023pdpp}              & 13.22&7.49\\
E3P $^\clubsuit$  \cite{wang2023event}              & 8.96&5.76\\ 
SkipPlan $^\clubsuit$  \cite{li2023skip}            & 8.55& 5.12\\\midrule 
Ours  ($R$=2)                          & 8.17 & 5.32\\
Ours $^\clubsuit$ ($R$=1)                             & \textbf{13.25}& 8.09\\ 
Ours $^\clubsuit$ ($R$=2)                             & 12.74& \textbf{9.23}\\ 
\midrule

PDPP $^\clubsuit$ {\textdagger}  \cite{wang2023pdpp} & 13.45& 8.41\\ 
Ours $^\clubsuit$ {\textdagger}  ($R$=1)         & \textbf{14.20}& \textbf{9.27}\\

\bottomrule

\end{tabular}
\caption{Success Rate ($SR^{\uparrow}$) comparison to existing baselines for CrossTask dataset under longer horizons}

\label{tab:longer-horizon} \vspace{-0.6cm}
\end{table}

%\subsection{Evaluation Protocol}
\noindent \textbf{Datasets and implementation Details:} In our evaluation, we employed datasets from three sources: CrossTask \cite{zhukov2019cross}, COIN \cite{tang2019coin}, and the Narrated Instructional Videos (NIV) \cite{alayrac2016unsupervised}. See section C.4 of supplementary material for details on datasets.
% The CrossTask dataset comprises 2,750 video clips, each representing one of 18 distinct tasks, and features an average of $7.6\pm4.4$ actions per clip. The COIN dataset is more extensive, including 11,827 videos across 180 tasks, with an average of $3.9\pm2.4$ actions per video. Lastly, the NIV dataset, though smaller in scale, includes 150 videos that capture 5 everyday tasks, with a higher density of actions, averaging $8.8\pm2.8$ (mean$\pm$std) actions per video. We randomly select 70\% data for training and
% 30\% for testing as previous work \cite{wang2023pdpp, bi2021procedure, zhao2022p3iv}.
All ablation studies and analyses were conducted on CrossTask. 
We use two Tesla A100 GPUs for all the experiments. We chose horizon  $T\in \{3,4,5,6\}$ and P$^2$KG ($R$=1) condition for implementation. In some cases, we incorporate P$^2$KG ($R$=2) and LLM conditions which are indicated in the respective tables. Throughout this study, the P$^2$KG ($R$=1) is employed with a batch size of 256, unless explicitly stated otherwise. 
More implementation details are available in the section C.2 of the supplementary material.

 \begin{figure}[t]
    \centering
    \includegraphics[width=\linewidth]{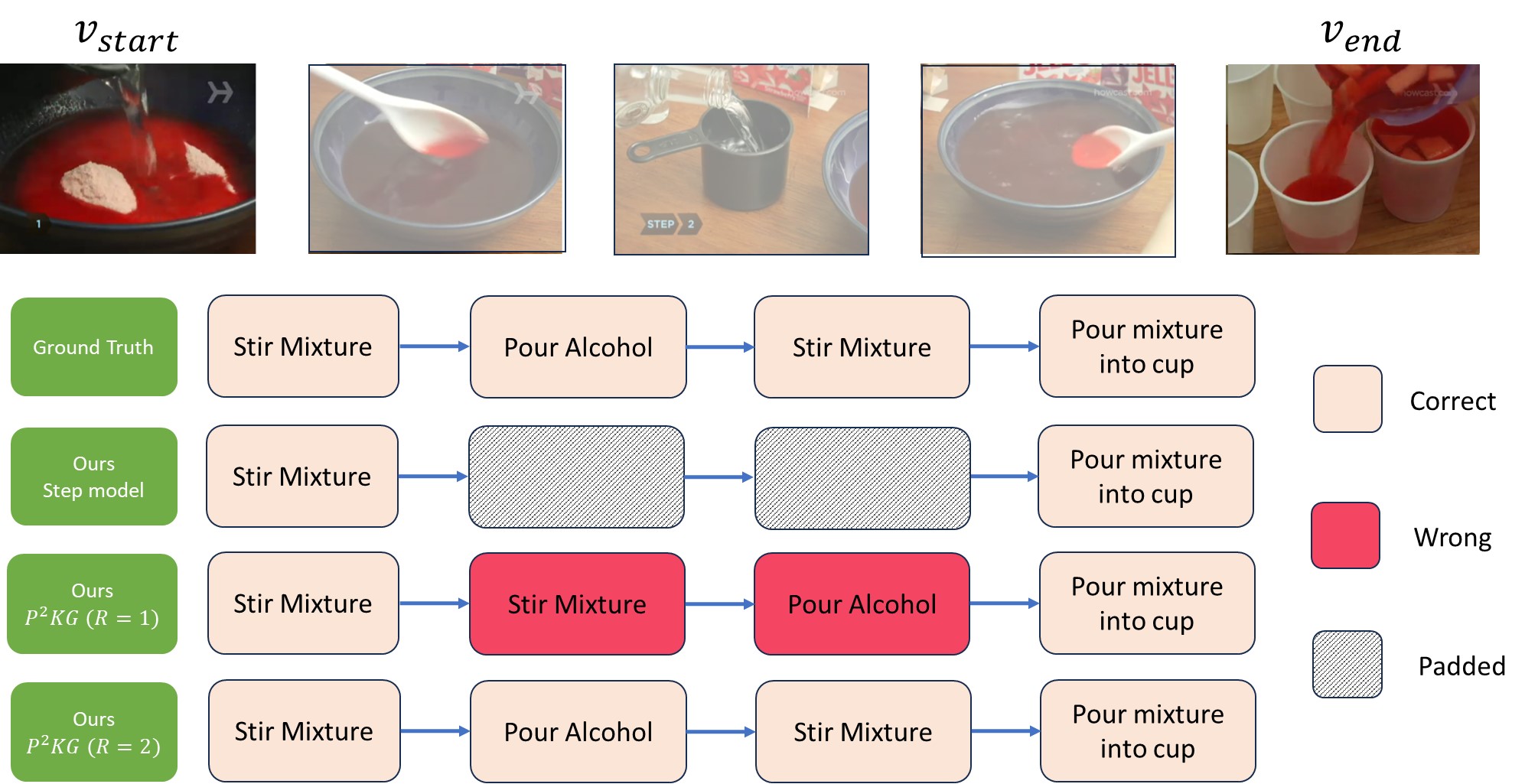}
    \caption{Qualitative analysis of the `Make Jello Shots' task}
    \label{fig:qualitative-incorrect}\vspace{-0.4cm}
\end{figure}

\begin{figure}[t]
    \centering
    \includegraphics[width=\linewidth]{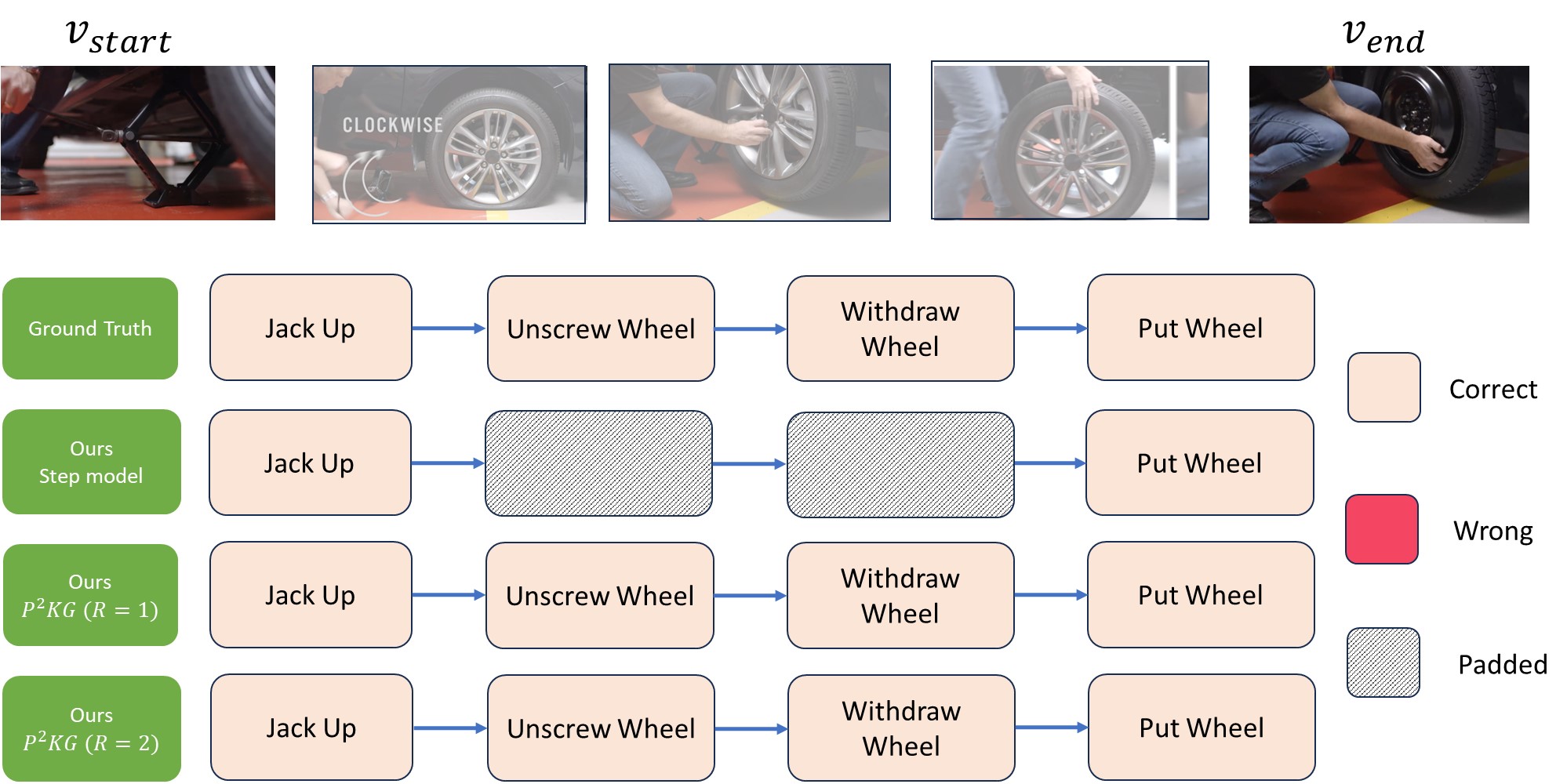}
    \caption{Qualitative analysis of the `Change a Tire' task}
    \label{fig:qualitative-correct} \vspace{-0.3cm}
\end{figure}

\noindent \textbf{Evaluation Metrics and baselines:} We use mean intersection over union (mIoU), mean accuracy (mAcc), and success rate (SR) as evaluation metrics. \textbf{SR is the most stringent metric}. See sec. C.4 of supplementary for more details.
%move to sup
% mIoU evaluates the overlap between
% the predicted actions with the ground-truth actions using IoU by regarding them as two action sets.
% mAcc is quantified as the average of the precise correspondences between the predicted actions and their respective ground truth counterparts at corresponding time steps. 
% Thus, mAcc counts the order of actions, and is stricter than mIoU.
% \textbf{SR is the most stringent metric}, considering a sequence to be successful solely if it exhibits a perfect match with the ground truth action sequence throughout.
We compare our model with 
%widely used 
state-of-the-art methods: WLTDO \cite{ehsani2018let}, UAAA \cite{abu2019uncertainty}, UPN \cite{srinivas2018universal},   DDN \cite{chang2020procedure}, PlaTe \cite{sun2022plate}, Ext-GAIL \cite{bi2021procedure}, P$^3$IV \cite{zhao2022p3iv}, PDPP \cite{wang2023pdpp}, SkipPlan \cite{li2023skip}, and E3P \cite{wang2023event}.
% and SCHEMA \cite{schema}. 
More details of these methods are available in the section C.5 of supplementary material.  Compared to other models, PDPP uses a different 
experimental 
setting. In PDPP, authors set the window after the start time of $a_1$ and before the end time of $a_T$, contrary to the standard practice of setting a
2-second
window around the start and end time (\textit{ref.}~\cite{chang2020procedure}).  
\iffalse
This approach could inadvertently incorporate elements of the action steps themselves, particularly in short-duration actions, thus skewing the results. 
\fi
We conduct experiments on both PDPP's proposed setting and the conventional setting. 

\noindent \textbf{Inference:} During the inference phase, the model receives only the start observation \( v_s \) and the goal observation \( v_g \). To proceed, it utilizes a step model to predict the initial action \( a_1 \) and the end action \( a_T \) for each data. Subsequently, leveraging the P$^2$KG, highest probable procedure knowledge graph plans connecting \( a_1 \) and \( a_T \) are obtained. Then, a multi-dimensional array, is created as mentioned in Eq.~\ref{eq:matrix-multi}. Finally, the planning model is used to predict the sequence of actions \( [a_1, ..., a_T] \) by denoising the generated multi-dimensional array as in~\S~\ref{subsubsec:planning-model-ref}. 

\subsection{Comparison with the State of the Art (SOTA)}

\noindent \textbf{CrossTask (short horizon)}: 
We evaluate on CrossTask for short horizons ({\( T=3 \)} and {\( T=4 \)}). 
According to the results shown in Table \ref{tab:cross-task}, our proposed method outperforms the PDPP in PDPP's setting in every evaluation metric. More than 0.9\% and 2\% improvement in success rate in \( T=3 \) and \( T=4 \) respectively. In the conventional setting,
our method with both P$^2$KG ($R$=1) and  P$^2$KG ($R$=2) conditions outperform the success rate values by a significant margin compared to other baselines. 
P$^2$KG ($R$=2) slightly outperforms P$^2$KG ($R$=1), indicating potential benefits of incorporating more procedural knowledge from the P$^2$KG.

\noindent \textbf{CrossTask (long horizon)}: 
We use long-horizon predictions for \( T=5 \) and \( T=6 \) for further evaluating our model as shown in Table \ref{tab:longer-horizon}. 
\iffalse
Here also we evaluate our model with CrossTask$_{How}$
for both  P$^2$KG ($R$=1) and  P$^2$KG ($R$=2) conditions in conventional settings and for with CrossTask$_{Base}$ for P$^2$KG ($R$=2) condition as shown in Table \ref{tab:longer-horizon}. 
\fi
In PDPP's setting ({\textdagger}), our method improves the success rate in both \( T=5 \) and \( T=6 \). In the conventional setting, our method utilizing  P$^2$KG ($R$=1) demonstrates the highest SR value for \( T=5 \), and for a longer horizon at \(T=6\), our method delivers superior performance for P$^2$KG ($R$=2).
Our method 
performs well under the challenging scenario of a long planning horizon.
Our success rate (SR) diminishes from approximately 40\% to 10\% when extending the planning horizon from T=3 to T=6, primarily due to the heightened uncertainty surrounding the predicted plan between the initial and final steps. This uncertainty stems from the increase in the number of potential procedural plans within the P$^2$KG.

\noindent \textbf{NIV and COIN}: 
% Evaluations of our method on other datasets 
Results are shown in 
% Table \ref{tab:niv-coin-results}.
Table \ref{tab:niv-results} and Table \ref{tab:coin-results}. 
% On the NIV dataset, 
On NIV, ours achieves the best result under the mIoU metric with $T$=3, and under both the SR and mIoU metrics with $T$=4. The results on NIV ($T$=5, $T$=6) are available in the \suppeos.
For the COIN dataset we only report SR and mAcc
% Success Rate (SR), and mean accuracy (mAcc) 
due to space constraints; mIoU is reported in the supplementary material. Our method does not rank as the top performer on COIN when $T$=3 or $T$=4. The likely reason is that the COIN dataset features just an average of 3.9 actions per video--a scenario that demands only short-horizon planning and does not necessitate \textit{advanced} procedural knowledge (which encompasses long sequence-level knowledge~\cite{zhou2023procedure}). Furthermore, the dataset's extensive collection of over 11k videos provides a substantial resource for baselines to learn \textit{basic} procedural knowledge.

\begin{table}[!htp]
\setlength{\tabcolsep}{3.6pt}
\footnotesize
  \aboverulesep=0ex
  \belowrulesep=0ex 
\centering
\begin{tabular}{@{}lccc|ccc@{}}
\toprule
 Models    & \multicolumn{3}{c|}{NIV ($T$=3)}      & \multicolumn{3}{c}{NIV ($T$=4)}      \\ \cline{2-7}
      &      \( SR^{\uparrow} \) & \( mAcc^{\uparrow} \) & \( mIoU^{\uparrow} \) &      \( SR^{\uparrow} \) & \( mAcc^{\uparrow} \) & \( mIoU^{\uparrow} \) \\ \midrule
   Random       & 2.21 & 4.07 & 6.09  & 1.12 & 2.73 & 5.84 \\ 
 DDN \cite{chang2020procedure}     & 18.41& 32.54& 56.56  & 15.97& 27.09& 53.84 \\
 Ext-GAIL \cite{bi2021procedure}   & 22.11& 42.20& 65.93 & 19.91& 36.31& 53.84    \\
 P$^3$IV  \cite{zhao2022p3iv}   & 24.68& 49.01& 74.29  & 20.14& 38.36& 67.29  \\  
 E3P~\cite{wang2023event}       &  \textbf{26.05} & \textbf{51.24} &  75.81  & 21.37 & \textbf{41.96} &  74.90   \\ 
 PDPP \cite{wang2023pdpp} & 22.22 & 39.50 & 86.66 & 21.30 & 39.24& 84.96  \\ 
 \midrule          
 Ours              & 24.44 & 43.46 & \textbf{86.67}    & \textbf{22.71} & 41.59 & \textbf{91.49}    \\ \bottomrule
\end{tabular}
\caption{Performance of baselines and ours for NIV dataset}
\label{tab:niv-results}
\end{table}

\vspace{-5pt}
\begin{table}[!htp]
\setlength{\tabcolsep}{3.6pt}
\footnotesize
  \aboverulesep=0ex
  \belowrulesep=0ex 
\centering
    \begin{tabular}{@{}lcc|cc|cc@{}}
\toprule
 Models    & \multicolumn{2}{c|}{COIN ($T$=3)}      & \multicolumn{2}{c|}{COIN ($T$=4)} & \multicolumn{2}{c}{COIN ($T$=5)}      \\ \cline{2-7}
      &      \( SR^{\uparrow} \) & \( mAcc^{\uparrow} \) &      
      \( SR^{\uparrow} \) & \( mAcc^{\uparrow} \) &     
      \( SR^{\uparrow} \) & \( mAcc^{\uparrow} \)  \\ \midrule
   Random       &  \( <0.01 \) & \( <0.01 \)
&\( <0.01 \) & \( <0.01 \) 
& - & -  \\ 
 Retrieval     &  4.38        & 17.40       
&2.71& 14.29
& - & -  \\
 DDN \cite{chang2020procedure}   &  13.90        & 20.19  
& 11.13& 17.71
& - & - \\

 P$^3$IV  \cite{zhao2022p3iv} &   15.40        & 21.67  
&11.32& 18.85
& 4.27& 10.81\\ 
    E3P~\cite{wang2023event}       &  19.57 & 31.42  
&13.59& 26.72& - & -  \\ 
 PDPP \cite{wang2023pdpp} & 19.42 & 43.44 & 
 13.67 &  42.58 & 
 13.02&  \textbf{43.36} \\

  SkipPlan~\cite{li2023skip}       & \textbf{23.65} & \textbf{47.12} 
& \textbf{16.04}& \textbf{43.19}& 9.90& 38.99 \\  \midrule     
 Ours ($R$=2)     & 20.25 & 39.87  & 15.63 & 39.53& \textbf{16.06} & 40.72 \\

 \bottomrule 
\end{tabular}
\caption{Performance of baselines and ours  
for COIN dataset  
}
\label{tab:coin-results}
\end{table}
% \fi

\begin{table*}
\vspace{-10pt}
   \setlength{\tabcolsep}{6.6pt}
%\small
\footnotesize
  \aboverulesep=0ex
  \belowrulesep=0ex 
\centering
    \begin{tabular}{lllllllllllll}
    \toprule
         Model&  \multicolumn{3}{c}{T=3}&  \multicolumn{3}{c}{T=4}&  \multicolumn{3}{c}{T=5} & \multicolumn{3}{c}{T=6}\\
          \cmidrule(l){2-4} \cmidrule(l){5-7} \cmidrule(l){8-10} \cmidrule(l){11-13}
         &  SR&  mAcc&  mIoU&  SR&  mAcc&  mIoU&  SR&  mAcc& mIoU & SR& mAcc&mIoU \\ \midrule
         w.o P$^2$KG conditions {\textdagger} &  35.69&  63.91&  66.04&  20.52&  57.47&  64.39&  12.8&  53.44&  64.01& 8.15& 50.45&64.13
\\
         Ours {\textdagger} &  38.12&  64.74&  67.15&  24.15&  59.05&  66.64&  14.20 &  53.84 & 65.56 & 9.27& 50.22&65.97\\ \midrule
         
         w.o P$^2$KG conditions &  31.35& 59.51& 63.11&  18.92&  56.20&  62.47&  
         12.71& 51.29&  63.56& 
         8.16& 47.63 &63.39
\\
         Ours  &  33.38 &  60.79 &  63.89& 
         21.02&  56.08&  64.15&  
         12.74 &  51.23 & 63.16&
         9.23& 50.78 & 65.56\\ \bottomrule
    \end{tabular}
    \caption{Performance of our method with and without P$^2$KG conditions on CrossTask $^\clubsuit$ dataset
    }
    \label{tab:w.o.conditions}
\end{table*}

\begin{table}[!h]
    \setlength{\tabcolsep}{10.6pt}
%\small
\footnotesize
  \aboverulesep=0ex
  \belowrulesep=0ex 
\centering
% \vspace{-10pt}
    \begin{tabular}{llll}
    \toprule
         Model ($T$=6, CrossTask $^\clubsuit$) &  SR&  mAcc& mIoU\\
         \midrule
         \multicolumn{4}{c}{Ours with P$^2$KG ($R$=1) }\\  \midrule
      \cellcolor{blue!15}   PDPP setting& \cellcolor{blue!15}  \textbf{9.27}&  \cellcolor{blue!15} 50.22& \cellcolor{blue!15} 
 \textbf{65.97}\\
         Conventional setting&  8.09&  50.80& \textbf{65.39}\\  \midrule
         \multicolumn{4}{c}{One LLM plan recommendation }  \\  \midrule
        \cellcolor{blue!15} PDPP setting (13b) & \cellcolor{blue!15} 7.74& \cellcolor{blue!15} 50.28& \cellcolor{blue!15} 64.05\\
         Conventional setting (13b)&  7.21&  49.68& 63.89\\
         \cellcolor{blue!15} PDPP setting (70b) & \cellcolor{blue!15} 8.62& \cellcolor{blue!15} \textbf{50.31}& \cellcolor{blue!15} 64.34\\
         Conventional setting (70b)&  7.81&  49.75& 64.02\\
         \midrule
         \multicolumn{4}{c}{P$^2$KG ($R$=1) and one LLM plan recommendation}\\  \midrule
         \cellcolor{blue!15} PDPP setting (13b) & \cellcolor{blue!15} 8.81& \cellcolor{blue!15} 49.97& \cellcolor{blue!15} 65.22\\
Conventional setting (13b)&  8.20&  51.46& 64.30\\
         \cellcolor{blue!15} PDPP setting (70b) & \cellcolor{blue!15} 9.01& \cellcolor{blue!15} 50.25 & \cellcolor{blue!15} 65.57\\
Conventional setting (70b)&  \textbf{8.34}&  \textbf{51.53}& 64.96\\
\bottomrule
    \end{tabular}
    \caption{Performance of the plan recommendations provided by the probabilistic procedure knowledge graph v.s. LLM.
    }
    \label{tab:pkg vs llm}
\end{table}

%\vspace{-1pt}
\subsection{Ablation Studies and Analyses}

\noindent \textbf{Ablation on the probabilistic procedure knowledge graph.}
We analyze the role of P$^2$KG in improving the performance of our proposed method. Table \ref{tab:w.o.conditions} shows the 
% evaluated
results 
\iffalse
for $ T \in \{3, 4, 5, 6\} $ in PDPP's setting. 
Results 
\fi
which clearly demonstrate that using P$^2$KG conditions improves the performance significantly for every \(T\) value. Especially when \(T=4\), success rate (SR) improves more than 3\% and mean IoU improves more than 2\%.   

\noindent \textbf{Plan recommendations provided by probabilistic procedure knowledge graph v.s. LLM.}
We recognize the recent trend of utilizing LLMs to enhance action anticipation~\cite{zhao2023antgpt} or planning in other realms~\cite{singh2023progprompt,ajay2023compositional,Patel_Desai_2023,huang2022language,huang2022inner}.
In Table \ref{tab:pkg vs llm}, we compare the results between using P$^2$KG  v.s. using LLM (`llama-2-13b-chat' and `llama-2-70b-chat') to generate the plan recommendations.
%By looking at Table \ref{tab:pkg vs llm}, we can confirm that  using the P$^2$KG recommendation gives more accurate results than just using the LLM generated recommendation. 
When examining Table \ref{tab:pkg vs llm}, it becomes apparent that there are trade-offs between using LLM-generated recommendations and P$^2$KG recommendations. For instance, P$^2$KG recommendations are constrained by the data available in the training set, limiting their applicability to unseen procedural activities. On the other hand, LLMs tend to exhibit better generalization to such unseen activities. However, considering that the training and testing are conducted on the aforementioned three datasets with known activities, 
% it is evident that
P$^2$KG recommendations can yield more accurate results compared to relying on LLM-generated recommendations.

\noindent \textbf{Probabilistic procedure knowledge graph (P$^2$KG) v.s. Frequency-based procedure knowledge graph (PKG).}
The probabilistic procedure knowledge graph uses out-edge normalization to encode step transition probabilities (\S~\ref{subsubsec:p2kg}), while the frequency-based procedure knowledge graph uses min-max normalization over the frequency counts throughout the graph. In both cases, the planning model only uses one procedure plan recommendation from the
% procedure knowledge
graph as condition in our experimental analysis. By looking at the results shown in Table \ref{tab:frequency-knowledge}, it is evident that the probabilistic procedure knowledge graph outperforms the frequency-based procedure knowledge graph.

\begin{table}[h]
\setlength{\tabcolsep}{5pt}
\footnotesize
  \aboverulesep=0ex
  \belowrulesep=0ex 
\centering
\footnotesize
\begin{tabular}{lccc} 
\toprule
Models & SR & mAcc & mIoU \\ 
\midrule

Frequency graph & 7.66 & 48.61 & 64.21 \\ 
Probabilistic graph & \textbf{8.09} & \textbf{50.80} & \textbf{65.40} \\ \bottomrule
\end{tabular}
\caption{Performance comparison between probabilistic procedure knowledge graph v.s. frequency-based procedure knowledge graph for $T$=6 on CrossTask $^\clubsuit$ dataset}
\label{tab:frequency-knowledge}
\end{table}

\begin{table}
    \centering
    %\small
    \footnotesize
  \aboverulesep=0ex
  \belowrulesep=0ex 
    \begin{tabular}{llll}
    \toprule
        Condition & SR & mAcc & mIoU \\ \midrule
         without GT data aug. & 38.12 & 64.74 & 67.15 \\
        with GT data aug. & 32.45 & 62.42  & 62.80 \\ \bottomrule
    \end{tabular}
    \caption{Effect of different input conditions for performance on CrossTask $^\clubsuit$ dataset ($T$=3) in PDPP's setting }
    \label{tab:predicted steps}
\end{table}

\begin{table}
    \setlength{\tabcolsep}{2.6pt}
%\small
\footnotesize
  \aboverulesep=0ex
  \belowrulesep=0ex 
\centering
    % \scalebox{0.77}{
    \begin{tabular}{lllllllll} 
    \toprule
         Models&  \multicolumn{2}{c}{T=3}&  \multicolumn{2}{c}{T=4}&  \multicolumn{2}{c}{T=5}&  \multicolumn{2}{c}{T=6}\\ \cmidrule(l){2-9} 
  
         &  $\hat{a}_{1}$ &  $\hat{a}_{T}$ &  $\hat{a}_{1}$ &  $\hat{a}_{T}$&  $\hat{a}_{1}$ &  $\hat{a}_{T}$&  $\hat{a}_{1}$ &  $\hat{a}_{T}$ \\
                \midrule
         Ours   &  53.69&  50.60&  55.56&  52.51&  55.58&  51.81& 57.09 & 51.92\\
         Ours $^\clubsuit$ &  71.42&  63.32&  72.98&  63.37&  72.42&  63.29& 63.82 & 59.96\\
         \bottomrule
    \end{tabular}
    % }
    \caption{The step model's start and end step prediction accuracies on the CrossTask dataset
    }
    \label{tab:steps} \vspace{-0.4cm}
\end{table}

\noindent \textbf{Effect of utilizing \textit{predicted} steps for input conditions to train the procedure planing model.}
Our proposed problem decomposition allows training the planning model with ground truth (GT) first and last steps.
We experiment with two ways to train the planning model. Method 1 uses the predicted start and end steps ($\hat{a}_1$ and $\hat{a}_T$) as input to generate P$^2$KG conditions and use them to train the planning model. Method 2 is where we augment the predicted start and end steps using the GT start and end steps ($a_1$ and $a_T$) by generating 3 more data samples 
as follows: [$\hat{a}_1$,$a_T$], [$a_{1}$,$\hat{a}_T$], and [$a_1$,$a_T$].
Then we generate P$^2$KG conditions for each data and train the model.   
From the results shown in Table \ref{tab:predicted steps},
the method without GT data augmentation shows better results.
This suggests that
leveraging ground truth data in training can lead to worse performance in testing.
% While the step model does not consistently yield flawless predictions, this experiment validates the utility of using its predicted steps to train the planning model within our proposed problem decomposition.

%\subsection{Analyses}

\noindent \textbf{Qualitative results}.
Figures \ref{fig:qualitative-incorrect} and  \ref{fig:qualitative-correct} provide qualitative examples of our method. Intermediate steps are padded in the step model because it only predicts the start and end actions. In the `make jello shot' task (see Figure \ref{fig:qualitative-incorrect}), the model gives a wrong prediction in the intermediate steps when using P$^2$KG ($R$=1) condition. However, it predicts correctly when using  P$^2$KG ($R$=2) conditions. In the `change a tire' task shown in Figure \ref{fig:qualitative-correct}, the model is able to predict all the intermediate steps in given conditions.

\begin{figure}
    \centering
    \vspace{-15pt}
    \includegraphics[width=\linewidth]{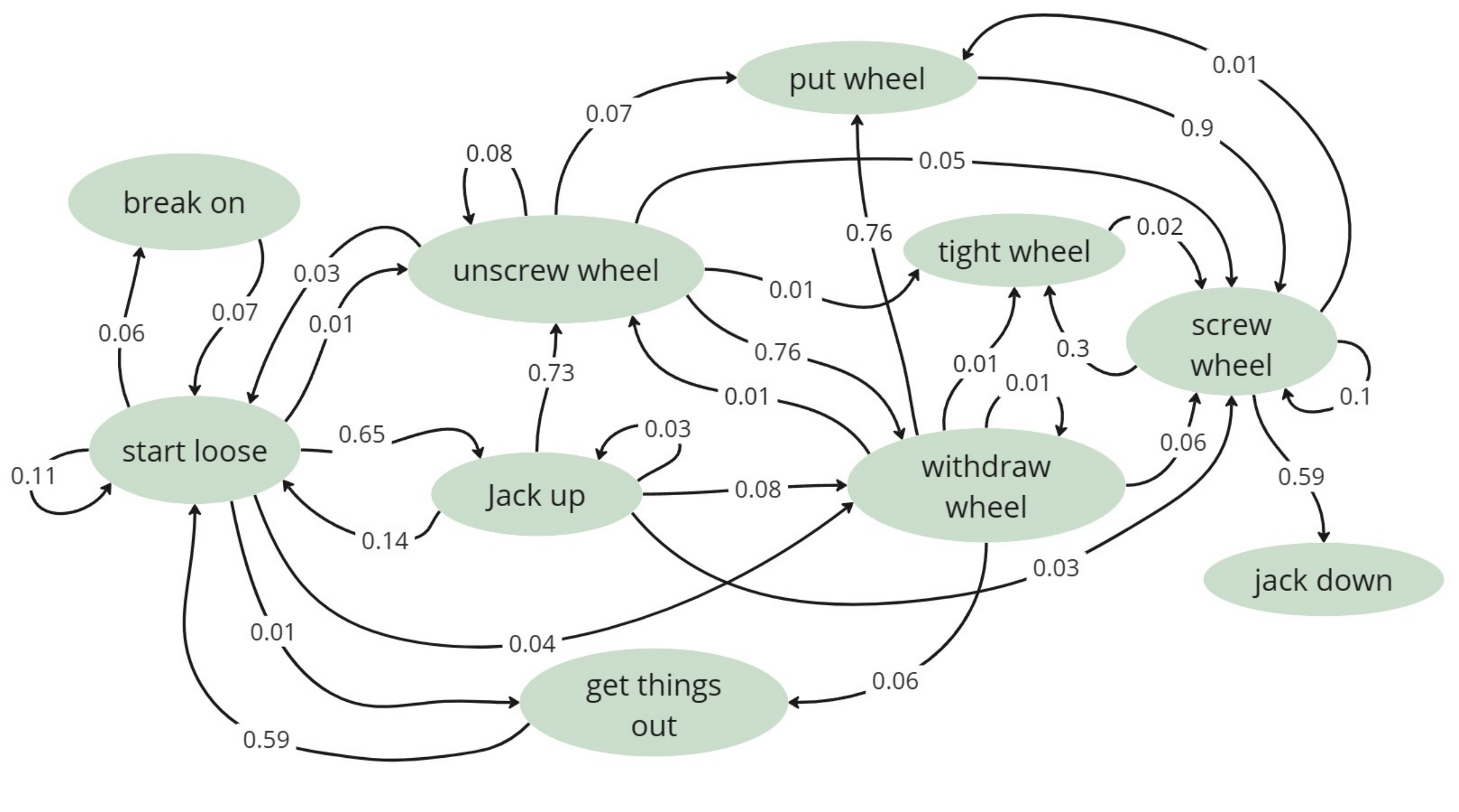}
    \vspace{-20pt}
    \caption{Example of a sub-graph in our probabilistic procedure knowledge graph (P$^2$KG) for CrossTask dataset. This graph effectively encapsulates real-world knowledge of distinct transition probabilities between steps, e.g., the probability of transitioning from `start loose' to `jack up' is 0.65, in contrast to a mere 0.14 for the reverse transition--the P$^2$KG reflects the common real-life practice where loosening the lug nuts before jacking up the car leads to a safer and more efficient tire change.}
    \label{fig:procedure_knowledge_graph}
\end{figure}

\begin{figure}
    \centering
    \vspace{-5pt}
\includegraphics[width=\linewidth]{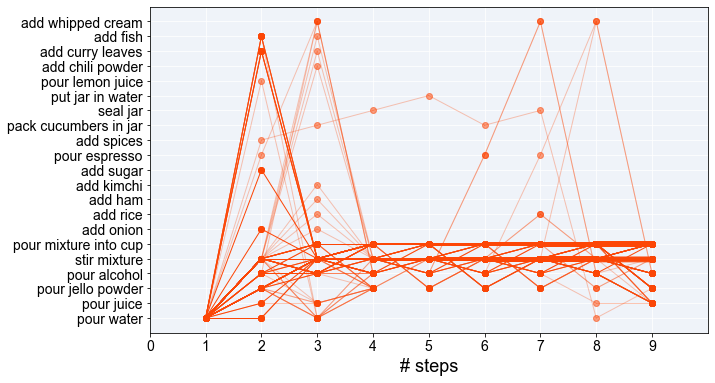}
\vspace{-20pt}
    \caption{Expert trajectories of the `Make Jello Shots' task,
    involving task-sharing steps and thus out-of-task step transitions. 
    Thicker lines indicate paths that are more frequently visited }
    \label{fig:expet-all} 
\end{figure}

\noindent \textbf{Visualizations of the probabilistic procedure knowledge graph}.
We show a sub-graph from our probabilistic procedure knowledge graph (Figure \ref{fig:procedure_knowledge_graph}).
This graph is drawn around the `jack up' node up to the depth of 2 nodes.

\noindent \textbf{Visualizations of the expert trajectories}.
Figure \ref{fig:expet-all} illustrates the steps involved in completing the `make jello shots' task, along with their transitions to other steps within the entire training data. This figure demonstrates that our P$^2$KG encodes diverse sequencing possibilities for steps and also captures task-sharing steps across the entire training domain. For instance, `pour water' is a step in `make jello shots' task, but it can also be part of other tasks, leading to a step transition from `pour water' to `add fish.' This structure allows models to leverage rich procedural knowledge.
\iffalse
Figure \ref{fig:expet-all} illustrates the steps involved in completing the `make jello shots' task. Thicker lines show more frequently visited paths. 
\fi

\noindent \textbf{Results for the step model}.
Table \ref{tab:steps} reveals step model results, indicating potential enhancement areas to elevate the planning performance.

\noindent \textbf{Limitations \& Failure cases}.
Our model exhibits three distinct failure case patterns. See section B.5 of the \supp for detailed discussions. 

\iffalse
\noindent \textbf{Failure cases}.
Our model exhibits three distinct failure case patterns: (a) when repetitive smaller action sequences exist within the procedure plan, (b) failure in accurately predicting $a_1$ and $a_T$ from the step model, and (c) failure in generating valuable P$^2$KG when $a_1$ and $a_T$ are the same. More details are provided in the section B.5 of the \suppeos. 
When comparing the error occurrences on CrossTask$^\clubsuit$ test data ($T$=4, $R$=2), $49.94\%$ of predicted plans have incorrect $\hat{a}_1$ or $\hat{a}_T$, and $63.23\%$ of error instances feature incorrect $\hat{a}_1$ or $\hat{a}_T$. 
If we use ground-truth $a_1$ and $a_T$ at inference, SR improves from $21.02$ into $36.58$. When analysing the errors of procedure knowledge graph (CrossTask$^\clubsuit$  ($T$=4, $R$=2) ), in $34\%$ of correctly predicted plans, the top-1 graph plan mismatches the ground-truth plan, and in $87\%$ of error instances, the top-1 graph plan mismatches the ground-truth plan. The analysis above implies that performance can be greatly boosted with more precise $\hat{a}_1$ and $\hat{a}_T$,
and having multiple plan recommendations is advantageous, as the top graph path may not always align.
\fi

%\vspace{-25pt}
\section{Conclusion}
\vspace{-5pt}

% We focus on the formulation of procedural plans from an AI agent in the realm of instructional videos. 
% We propose to enhance the agent's capabilities by incorporating procedural knowledge. This knowledge is gleaned from the training procedure plans,
% enabling the agent to adeptly handle the intricacies of step sequencing and its variations. We term this innovative approach \ours (Knowledge-Enhanced Procedure Planning). 
% \ours employs a probabilistic procedural knowledge graph, sourced from the training domain, effectively serving as a `textbook' for procedure planning.
% Experiments conducted on three datasets 
% demonstrate that \ours delivers top-tier performance while necessitating only a minimal amount of supervision. Future work can focus on enhancing step prediction accuracies to generate more precise plans by incorporating various image-to-text prediction algorithms for the initial and final steps.

We focus on formulating procedural plans from an AI agent in instructional videos. 
We propose 
% to enhance the agent's capabilities by incorporating procedural knowledge. We propose \ours in which this knowledge is gleaned from the training procedure plans,
% enabling the agent to adeptly handle the intricacies of step sequencing and its variations.
\ours which employs a probabilistic procedural knowledge graph, sourced from the training domain, effectively serving as a `textbook' for procedure planning.
Results 
show that \ours delivers SOTA performance with minimal supervision. 
Future work can focus on enhancing the accuracy of predictions for the initial and final steps. Additionally, our approach can be modified to aid in detecting erroneous steps and the misordering of steps in instructional videos~\cite{sener2022assembly101,narasimhan2023learning}.
% % step
% prediction accuracies
% % to generate more precise plans by incorporating various image-to-text prediction algorithms 
% for the initial and final steps. Our method can also be adapted to benefit the task of identifying mistake steps and misordering of steps in instructional videos~\cite{sener2022assembly101,narasimhan2023learning}.

{
    \small
    \bibliographystyle{ieeenat_fullname}
    \bibliography{main}
}

\clearpage
\setcounter{page}{1}
\maketitlesupplementary

\noindent Unless otherwise mentioned, all the results and analysis are obtained for the CrossTask dataset with input visual features from the S3D network \cite{miech2020end} pretrained on HowTo100M \cite{miech2019howto100m}. We organize the Supplementary Materials as follows:

\bigskip

%%%%%%%%%%%%%% 
\noindent \textbf{A. Further Experimental Results} 

\quad \textbf{A.1 Additional Results on COIN and NIV}

\quad \textbf{A.2 Parameter Sensitivity Analysis}

\noindent \textbf{B. More Thorough Analysis} 

\quad \textbf{B.1 More Visualizations of the P$^2$KG}

\quad \textbf{B.2 More Qualitative Results}

\quad \textbf{B.3 Training Efficiency}

\quad \textbf{B.4 Analysis by the Step Transition Heatmap}

\quad \textbf{B.5 Limitations and Failure Cases}

\quad \textbf{B.6 Ablations with Flawless Step Model}

\quad \textbf{B.7 Zero-shot Planning with the P$^2$KG} 

\quad \textbf{B.8 Further Discussions}

\noindent \textbf{C. Method and Implementation Details}

\quad \textbf{C.1 Diffusion Model Details}

\quad \textbf{C.2 Implementation of \ours}

\quad \textbf{C.3 Implementation of Ablations}

\quad \textbf{C.4 Datasets and Evaluation Metrics}

\quad \textbf{C.5 Baselines}

\bigskip
\bigskip

\section*{A. Further Experimental Results}
\subsection*{A.1 Additional Results on COIN and NIV}
\noindent \textbf{COIN.} We present the full results of our method in comparison to several baseline approaches on the COIN dataset across different planning horizons in Table \ref{tab:coin-all}. While the results demonstrate that our method outperforms the majority of previous literature, it does not secure the top position when considering small planning horizons ($T$=3 or $T$=4) for the COIN dataset. This can be attributed to the fact that the COIN dataset typically consists of only 3.9 actions per video on average, a scenario where advanced sequence-level procedural knowledge is not essential. 
The advantage of our method becomes increasingly pronounced as the planning horizon expands.
With a larger planning horizon ($T$=5), our method significantly outperforms the previous state-of-the-art (SOTA) methods, achieving a performance gain of $6.16$ on the most strict metric, Success Rate (SR), compared to SkipPlan~\cite{li2023skip}.
Our method's utilization of plan recommendations from the Probabilistic Procedure Knowledge Graph (P$^2$KG) effectively reduces the complexity of long-horizon planning.

\noindent \textbf{NIV.} Previous works did not provide results for the NIV dataset concerning long-horizon procedural planning; in Table~\ref{tab:niv_all}. we evaluate our model's performance in comparison to the PDPP model~\cite{wang2023pdpp} specifically on the NIV dataset for long-horizon procedural planning ($T$=5 or $T$=6). (For results regarding short-horizon planning, please refer to Table 3 in the main paper.) Our model demonstrates superior performance in terms of the SR, mAcc, and mIoU metrics.

\begin{table*}[!htp]
\renewcommand\thetable{S.1} 
  \aboverulesep=0ex
  \belowrulesep=0ex 
\centering
\begin{tabular}{@{}llllcccccc@{}}
\toprule
 Models    &  \multicolumn{3}{c}{COIN ($T$=3)}&\multicolumn{3}{c}{COIN ($T$=4)}      & \multicolumn{3}{c}{COIN ($T$=5)}      \\ \cline{2-10}
      &       \( SR^{\uparrow} \) & \( mAcc^{\uparrow} \) & \( mIoU^{\uparrow} \) &\( SR^{\uparrow} \) & \( mAcc^{\uparrow} \) & \( mIoU^{\uparrow} \) &      \( SR^{\uparrow} \) & \( mAcc^{\uparrow} \) & \( mIoU^{\uparrow} \) \\ \midrule
   Random       &  \( <0.01 \) & \( <0.01 \) & 2.47 
&\( <0.01 \) & \( <0.01 \) & 2.32
& - & - & - \\ 
 Retrieval     &  4.38        & 17.40       & 32.06 
&2.71& 14.29& 36.97
& - & - & - \\
 DDN \cite{chang2020procedure}   &  13.90        & 20.19       & 64.78 
&
11.13& 17.71& 68.06
& - & - & - \\

 P$^3$IV  \cite{zhao2022p3iv} &   15.40        & 21.67       & 76.31 
&11.32& 18.85&  70.53
& 4.27& 10.81&  68.81\\      
 E3P~\cite{wang2023event}       &  19.57 & 31.42 & 84.95    
&13.59& 26.72& 84.72& - & - & - \\ 
  PDPP \cite{wang2023pdpp} & 19.42 & 43.44 & 50.03 &
 13.67 &  42.58 & 49.84 &
 13.02&  \textbf{43.36} &  50.96 \\ 
 SkipPlan~\cite{li2023skip}       & \textbf{23.65} & \textbf{47.12} & \textbf{78.44} 
& \textbf{16.04}& \textbf{43.19}& \textbf{77.07}& 9.90& 38.99&\textbf{76.93}\\  \midrule     
 Ours ($R$=2)     & 20.25 & 39.87 & 51.72 & 15.63 & 39.53& 53.27& \textbf{16.06} &40.72 &56.15\\ \bottomrule
\end{tabular}
\caption{\textbf{Performance of baselines and ours on the COIN dataset.} Our method excels at handling challenging planning scenarios that demand longer planning horizons ($T$); with $T$=5, ours achieves a performance gain of 6.16 on the most strict metric, Success Rate (SR)
}
\label{tab:coin-all}
\end{table*}

\begin{table}
\renewcommand\thetable{S.2} 
\setlength{\tabcolsep}{2.4pt}
% \footnotesize
  \aboverulesep=0ex
  \belowrulesep=0ex 
    \centering
    % \small
    % \scalebox{0.9}{
    \begin{tabular}{lcccccc}
    \toprule
        Models &  \multicolumn{3}{c}{NIV ($T$=5)}&  \multicolumn{3}{c}{NIV ($T$=6)}\\
         &   \( SR^{\uparrow} \) & \( mAcc^{\uparrow} \) & \( mIoU^{\uparrow} \) &\( SR^{\uparrow} \) & 
         \( mAcc^{\uparrow} \) & \( mIoU^{\uparrow} \) \\ \cline{2-7}
         PDPP~\cite{wang2023pdpp}& 18.95 & 37.26 & 87.50 &  14.94 & 41.02 & 93.70\\
         Ours ($R$=2)& \textbf{21.58} &  \textbf{39.79} &\textbf{91.66}  & \textbf{17.53} & \textbf{43.62} & \textbf{93.75} \\
         \bottomrule
    \end{tabular}
    % }
    \caption{\textbf{Performance of baselines and ours on the NIV dataset.} Our method demonstrates superior performance on all metrics}
    \label{tab:niv_all}
\end{table}

\begin{table}
\renewcommand\thetable{S.3} 
% \setlength{\tabcolsep}{3.6pt}
% \footnotesize
  \aboverulesep=0ex
  \belowrulesep=0ex 
    \centering
    \begin{tabular}{cccc}
    \toprule
         Model&  \( SR^{\uparrow} \) &  \( mAcc^{\uparrow} \) & \( mIoU^{\uparrow} \) \\ \midrule
         PDPP~\cite{wang2023pdpp}& 18.69
  & 52.44 & 62.38\\
         Ours ($R$=1)& 20.38 & 55.54 &  64.03\\
         Ours ($R$=2)& \textbf{21.02} &  56.08 & \textbf{64.25} %64.15 
         \\
         Ours ($R$=3)& 20.22 & \textbf{56.19} & 63.15\\
         Ours ($R$=4)& 20.76 & 55.63 & 64.22 \\
         Ours ($R$=5)&  20.37&  55.43& 63.93 \\ \bottomrule
    \end{tabular}
    \caption{\textbf{Parameter sensitivity study on the CrossTask dataset.} Performance of our method initially increases and then decreases as $R$ continues to increase. An excessively large value for $R$ may lead to the inclusion of less prominent paths in the conditions given to the planning model}
    \label{tab:parameter-sensitivity}
\end{table}

\begin{figure*}[]
\renewcommand\thefigure{S.1} 
    \centering
    \begin{subfigure}[b]{0.48\textwidth}
        \includegraphics[width=\textwidth]{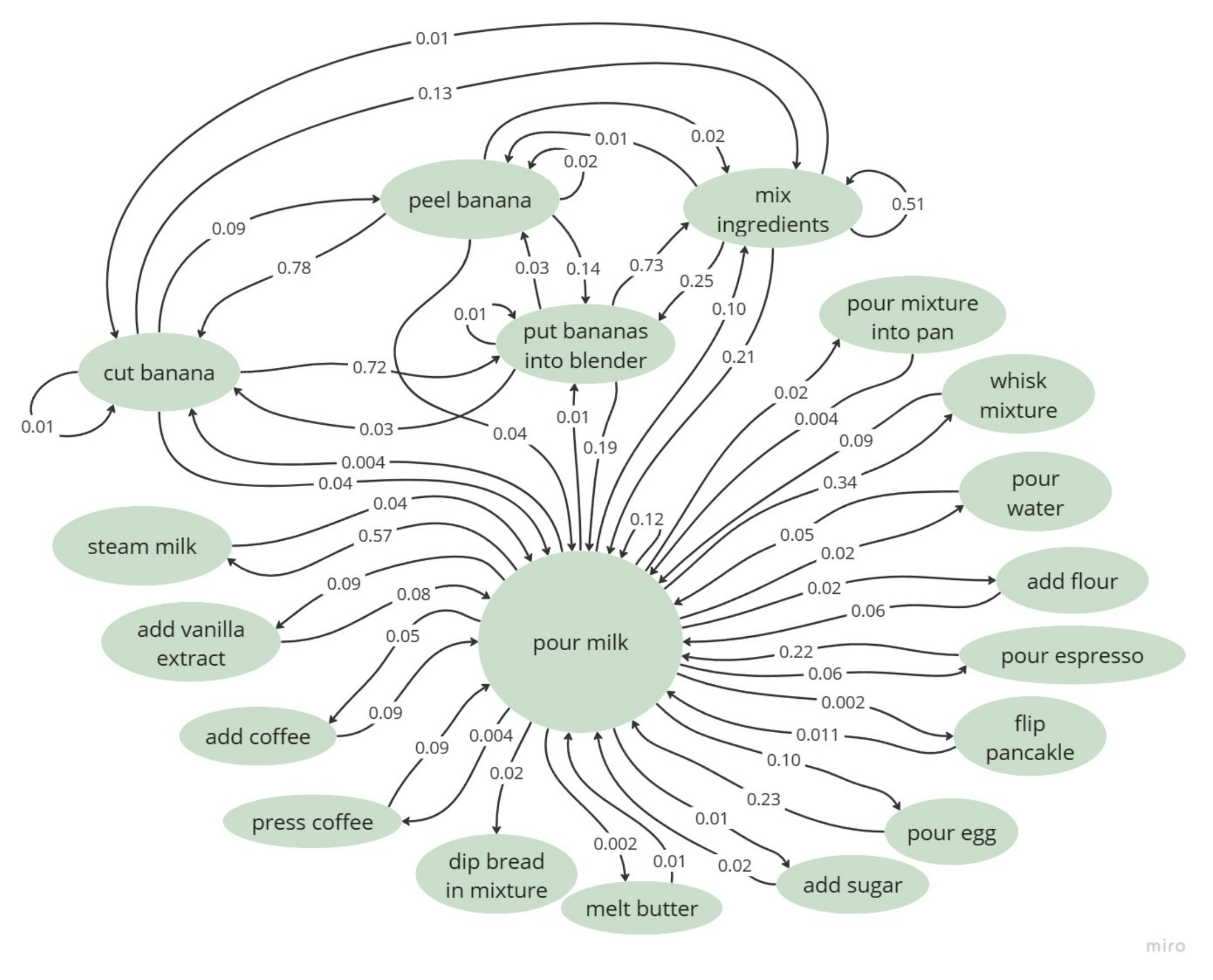}
        \caption{}
        \label{fig:image1}
    \end{subfigure}
    \hfill
    \begin{subfigure}[b]{0.48\textwidth}
        \includegraphics[width=\textwidth]{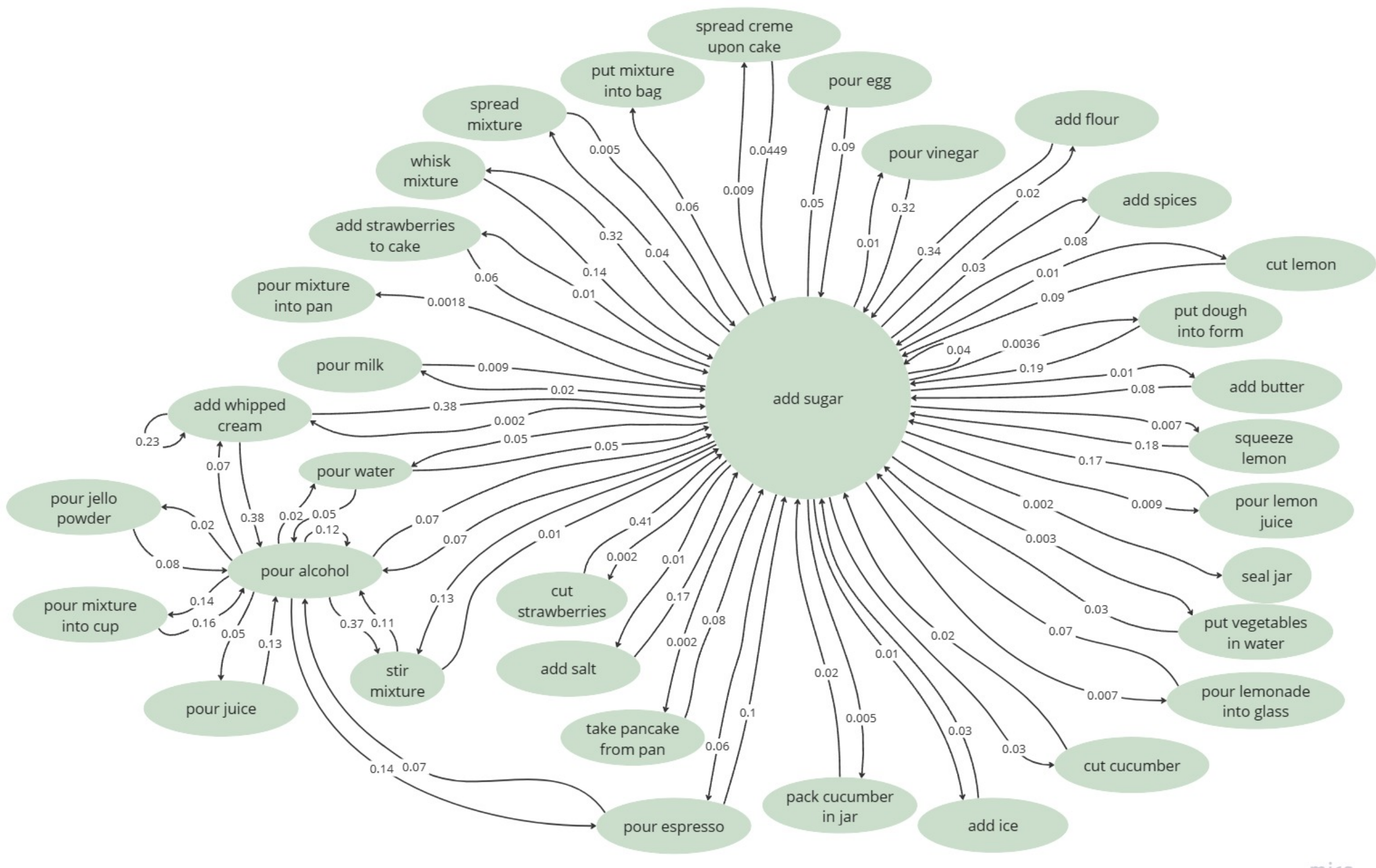}
        \caption{}
        \label{fig:image2}
    \end{subfigure}
    \caption{\textbf{Probabilistic Procedure Knowledge Graphs} (P$^2$KG) around the nodes `peel banana' and `add whipped cream' respectively up to a two-node depth. P$^2$KG effectively captures the task-sharing steps, variability in transition probabilities between steps, implicit temporal and causal relationships of steps, as well as the existence of numerous viable plans given an initial step and an end step
 }
    \label{pkg-examples-1}
\end{figure*}

\begin{figure*}
\renewcommand\thefigure{S.2} 
\centering
\begin{subfigure}{.5\textwidth}
  \centering
  \includegraphics[width=.9\linewidth]{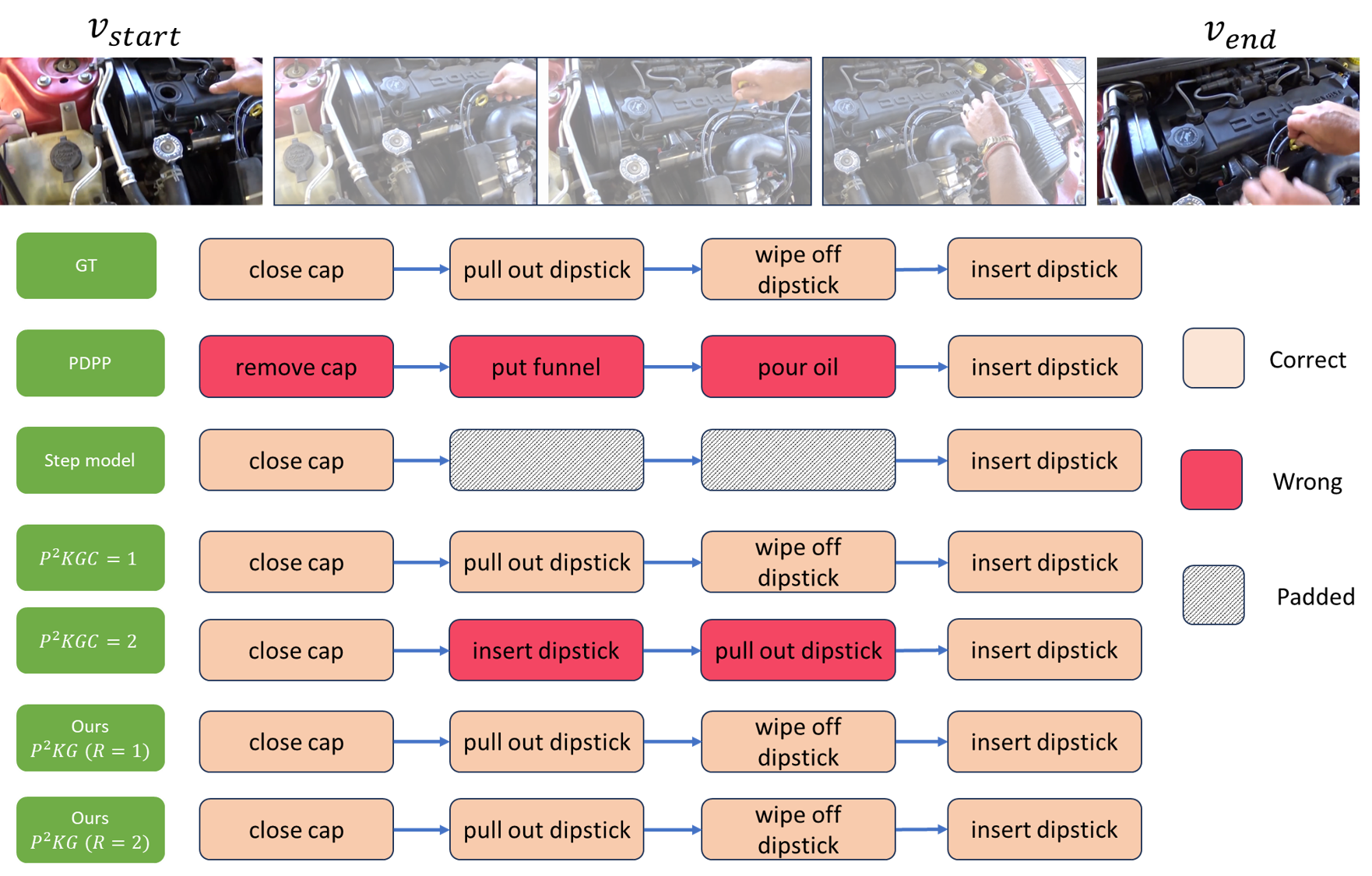}
  \caption{Add Oil to Your Car task}
  \label{fig:sub1}
\end{subfigure}%
\begin{subfigure}{.5\textwidth}
  \centering
  \includegraphics[width=.9\linewidth]{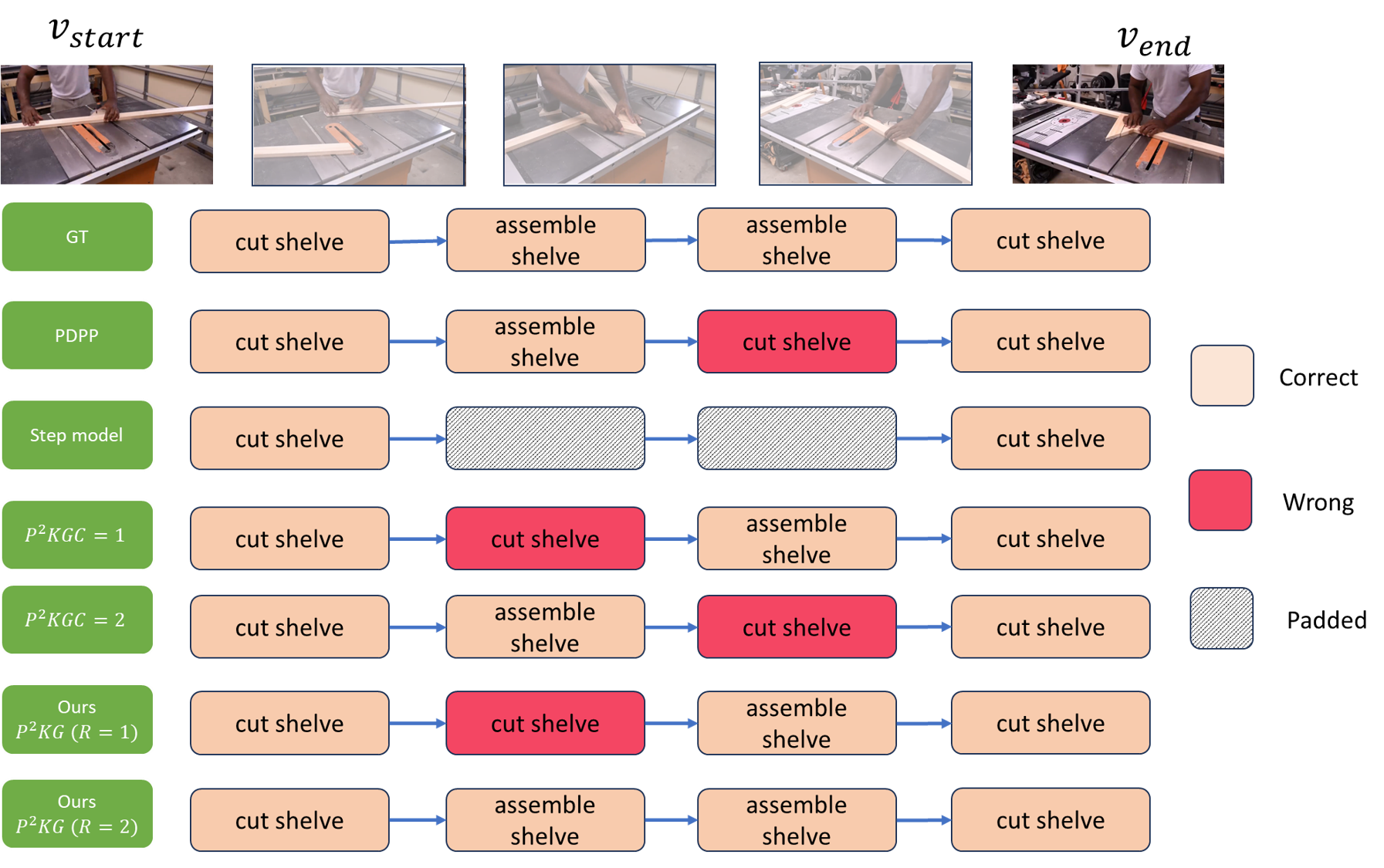}
  \caption{Build Simple Floating Shelves task}
  \label{fig:sub2}
\end{subfigure}
\newline
\begin{subfigure}{.5\textwidth}
  \centering
  \includegraphics[width=.9\linewidth]{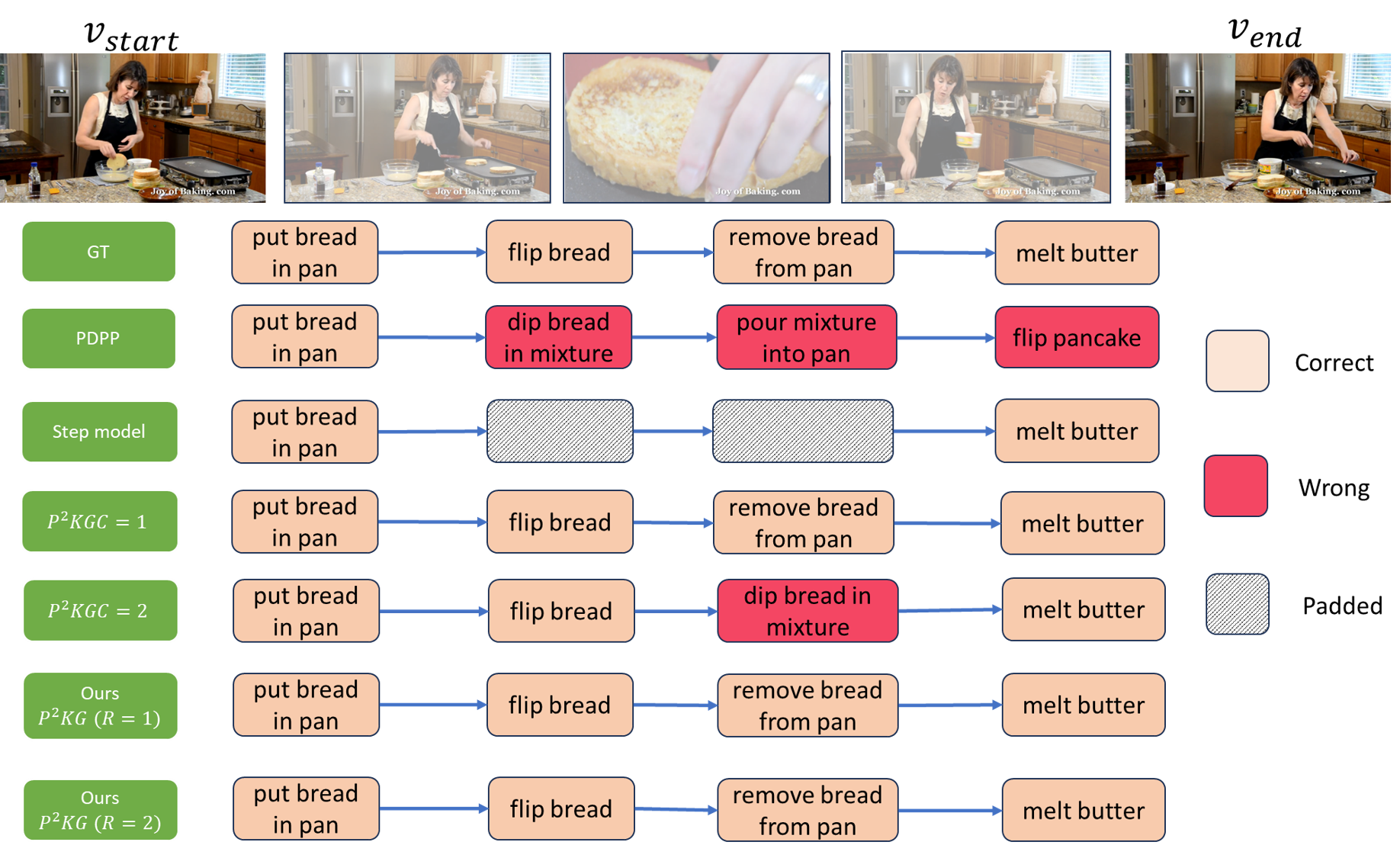}
  \caption{Make French Toast task}
  \label{fig:sub3}
\end{subfigure}%
\begin{subfigure}{.5\textwidth}
  \centering
  \includegraphics[width=.9\linewidth]{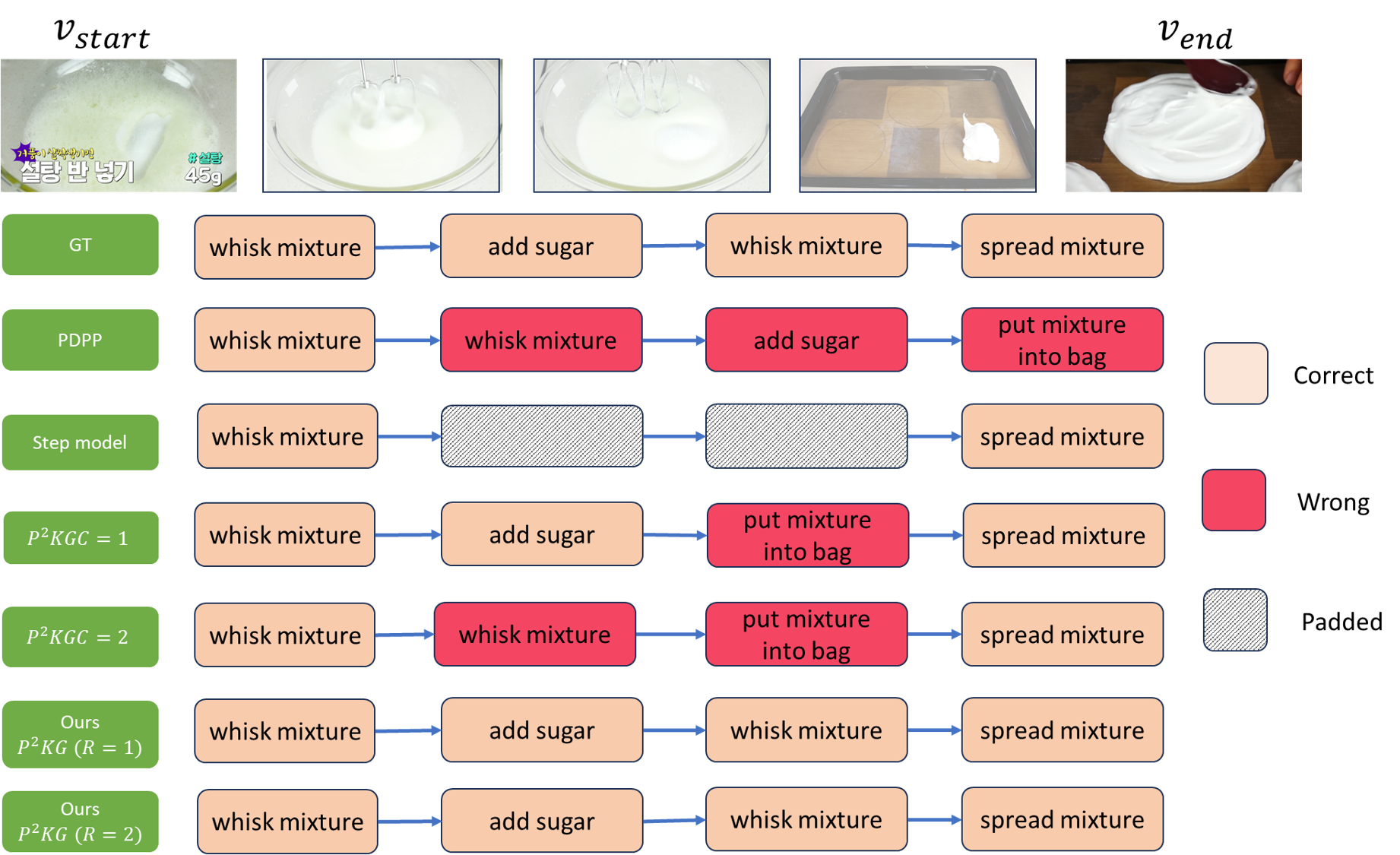}
  \caption{Make Meringue task}
  \label{fig:sub4}
\end{subfigure}
\caption{\textbf{Qualitative Analysis of Different Procedural Tasks.} In the illustration, `P$^2$KGC =1' and `P$^2$KGC =2' indicates the first and second paths obtained from the probabilistic procedure knowledge graph respectively }
\label{fig:test}
\end{figure*}

\subsection*{A.2 Parameter Sensitivity Analysis on the Number of Plan Recommendations} \label{subsubsec:param_sensitivity}
Our proposed method involves querying a probabilistic procedure knowledge graph; this process extracts the most probable graph paths, which have specified start and end steps. These paths are subsequently considered as recommended procedural plans and are employed as additional input conditions for the model, with the aim of improving its overall performance in procedure planning. 
Consequently, the number of plan recommendations, denoted as $R$ in the main paper, emerges as a novel hyper-parameter in our methodology.

We have conducted a parameter sensitivity analysis to examine the effect of the number of paths
selected from the probabilistic procedure knowledge graph 
to be given as conditions to the planning model on the procedure planning performance. Table \ref{tab:parameter-sensitivity} displays the parameter sensitivity of our method with respect to $R$ on the CrossTask dataset for the planning horizon $T$=4.  
To maintain simplicity in implementation, in cases where $R$ exceeds one, we aggregate the features of the top $R$ graph paths through a linear weighting process (i.e., weighted summation).
This results in consistent feature dimensions compared to when $R$=1. 
The weighting scheme is as follows:\newline
\fontsize{8}{12}\selectfont
$ 
R = 1 \xrightarrow{} weights: 1\newline
R =2 \xrightarrow{} weights: \frac{2}{3}, \frac{1}{3}\newline
R=3 \xrightarrow{} weights: \frac{3}{5}, \frac{1}{5}, \frac{1}{5} \newline
R=4 \xrightarrow{} weights: \frac{4}{7}, \frac{1}{7}, \frac{1}{7}, \frac{1}{7}\newline
R=5 \xrightarrow{} weights: \frac{5}{9}, \frac{1}{9}, \frac{1}{9}, \frac{1}{9}, \frac{1}{9}\newline
$
\normalsize
where the weights of the graph paths, starting with the most probable one, are provided above. The weights are empirically determined to emphasize the top one probable path by assigning it greater weight, and equal weights are distributed among the remaining graph paths. 

Our method, with $R$=2, achieves the best results in SR and mIoU (see Table \ref{tab:parameter-sensitivity}). In general, performance of our method initially increases and then decreases as $R$ continues to increase. The reason for larger values of $R$ yielding lower SR is attributed to the influence of less prominent paths being included in the conditions given to the planning model. 
Further tuning of the weighting scheme could potentially yield even more superior results
which we leave for future work.

Finally, a consistent trend of higher performance of $R=2$ over $R=1$  as $T$ increases is not always guaranteed. In Table 1 and Table 2 of the main paper, \ours with $R$=2 has a weaker performance than $R=1$ when $T$ is 5, but stronger than $R=1$ when $T$ is 3, 4 or 6.
This may arise from the stochastic nature of the diffusion model and the variability in the extent to which the top-2 graph plans offer additional beneficial information across different values of $T$.

\section*{B. More Thorough Analysis}

\subsection*{B.1 More Visualizations of the Probabilistic Procedure Knowledge Graph}
Figure \ref{pkg-examples-1} (a) and (b) show the sub-graphs in our probabilistic procedure knowledge graph (P$^2$KG) around the nodes `peel banana' and `add whipped cream' up to a two-node depth respectively. 
P$^2$KG effectively captures the task-sharing steps, variability in transition probabilities between steps, implicit temporal and causal relationships of steps, as well as the existence of numerous viable plans given an initial step and an end step.

\subsection*{B.2 More Qualitative Results}
Figure~\ref{fig:test} presents more qualitative results of our method compared with PDPP~\cite{wang2023pdpp} across a range of procedural tasks, with a planning horizon set at  $T$=4. In scenarios depicted in Figure~\ref{fig:test} (a), (c), and (d), the PDPP model erroneously predicts the initial or final steps based on the given visual states, resulting in procedural planning failures. Our approach, however, incorporates a specialized step model designed to accurately predict these crucial first and last steps. Furthermore, Figure~\ref{fig:test} highlights how our method benefits from the plan recommendations provided by the probabilistic procedure knowledge graph. These recommendations offer invaluable insights into feasible plans identified during training, thereby significantly enhancing our method's capability in procedure planning at test time.

\begin{figure*}
\renewcommand\thefigure{S.3} 
    \centering
    \includegraphics[width=1\linewidth]{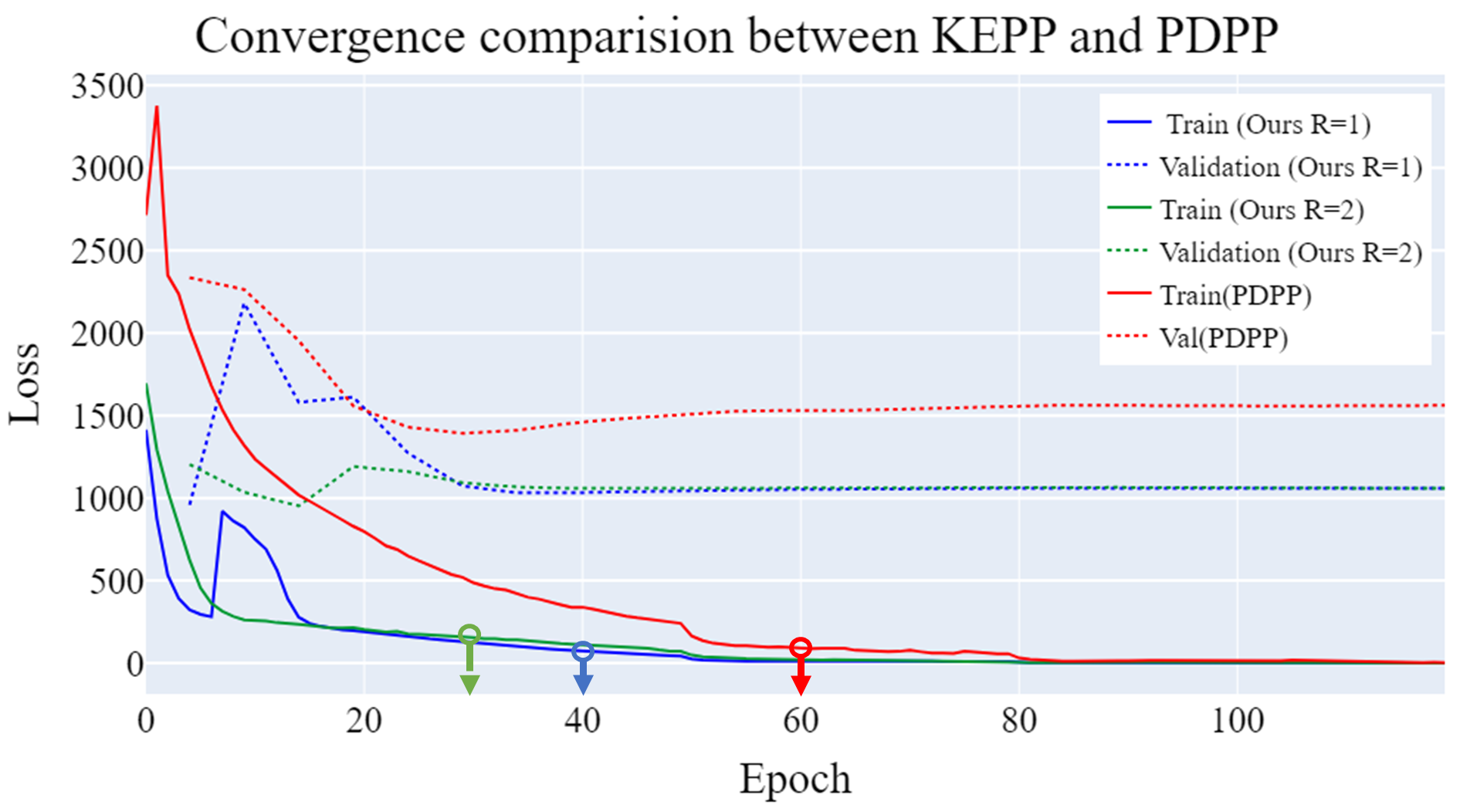}
    \caption{\textbf{Training efficiency comparison between PDPP~\cite{wang2023pdpp} and our model (\ourseos).} These training and validation profiles demonstrate that our model exhibited significantly faster training convergence compared to PDPP. Our method with $R$=2 has faster convergence compared to $R$=1. The arrows indicate the epochs of convergence determined by the early stopping scheme }
    \label{fig:convergence}
\end{figure*}

\subsection*{B.3 Training Efficiency}

Figure \ref{fig:convergence} compares the training convergence of our planning model with PDPP \cite{wang2023pdpp}. We have plotted the loss values across the training epochs and indicated the epoch of convergence determined by the early stopping scheme. Our model, with $R$=2 and $R$=1, converged at the 30th and 40th epochs, respectively, while the previous state-of-the-art method, PDPP, converged at the 60th epoch. Our model exhibited significantly faster training convergence compared to PDPP. The training efficiency of our method is achieved by leveraging the probabilistic procedure knowledge graph, which provides valuable context to facilitate the model's learning process.

\iffalse
\begin{figure*}[htbp]
    \centering
    \begin{subfigure}[b]{0.24\textwidth}
        \includegraphics[width=\textwidth]{fig/train.png}
        \caption{Train}
        \label{fig:image1}
    \end{subfigure}
    \hfill
    \begin{subfigure}[b]{0.24\textwidth}
        \includegraphics[width=\textwidth]{fig/test.png}
        \caption{Test}
        \label{fig:image2}
    \end{subfigure}
    \hfill
    \begin{subfigure}[b]{0.24\textwidth}
        \includegraphics[width=\textwidth]{fig/pdpp.png}
        \caption{PDPP}
        \label{fig:image3}
    \end{subfigure}
    \hfill
    \begin{subfigure}[b]{0.24\textwidth}
        \includegraphics[width=\textwidth]{fig/ours.png}
        \caption{Ours (R=1)}
        \label{fig:image4}
    \end{subfigure}
    \caption{Heat Map comparison for Change a Tire task in CrossTask dataset}
    \label{fig:four_images}
\end{figure*}
\fi

\begin{figure*}[htbp]
\renewcommand\thefigure{S.4} 
    \centering
    \vspace{10pt}
    \begin{subfigure}[b]{0.33\textwidth}
        \includegraphics[width=\textwidth]{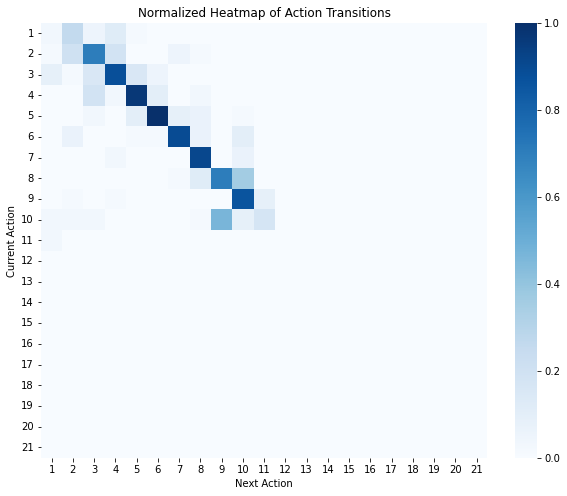}
        \caption{Train}
        \label{fig:image1}
    \end{subfigure}
    \hfill
    \begin{subfigure}[b]{0.33\textwidth}
        \includegraphics[width=\textwidth]{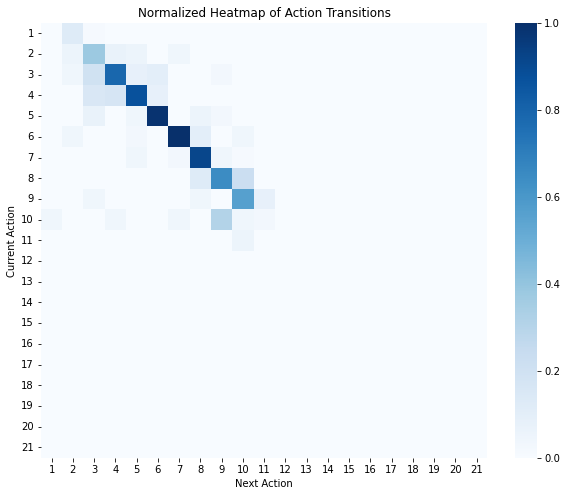}
        \caption{Test}
        \label{fig:image2}
    \end{subfigure}
    \hfill
    \begin{subfigure}[b]{0.33\textwidth}
        \includegraphics[width=\textwidth]{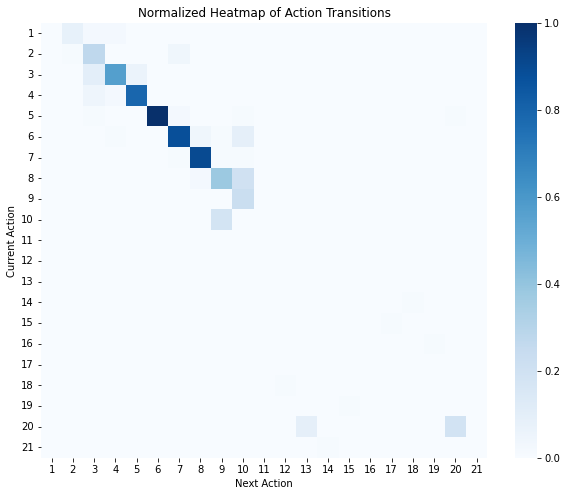}
        \caption{Ours (R=1)}
        \label{fig:image4}
    \end{subfigure}
    \caption{\textbf{Step Transition Heatmaps on the CrossTask dataset of (a) the training set, (b) the testing set, and (c) procedure plan predictions made by our method on the test set}. The i-row-j-column depicts the probability of the transition from i-th action to j-th action. Darker color indicates higher probability. Please refer to Section B.4 for details regarding task and action names}
    \label{fig:heatmap}
\end{figure*}

% \iffalse
\begin{figure*}[htbp]
\renewcommand\thefigure{S.5} 
    \centering
    \begin{subfigure}[b]{0.33\textwidth}
        \includegraphics[width=\textwidth]{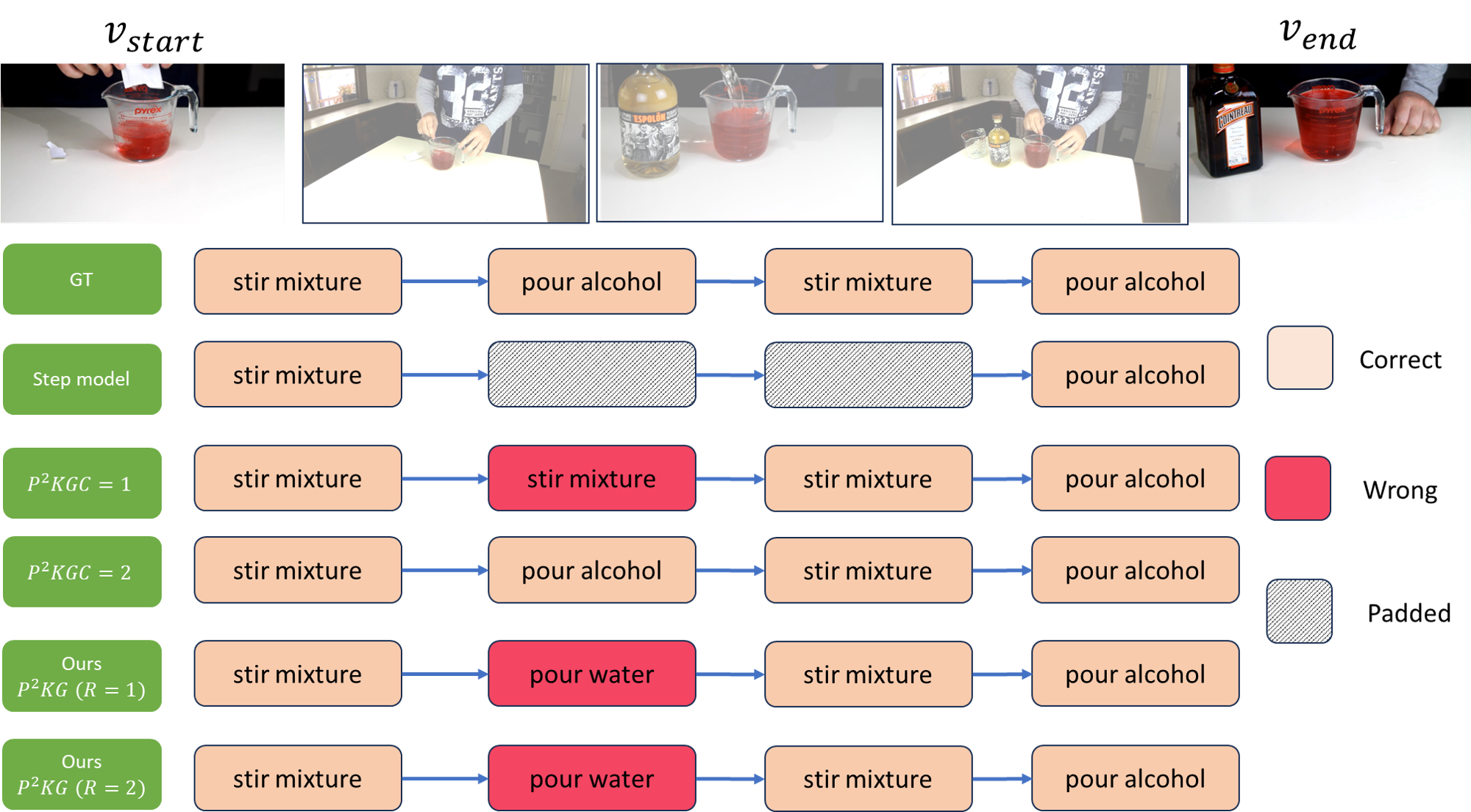}
        \caption{}
        \label{fig:image1}
    \end{subfigure}
    \hfill
    \begin{subfigure}[b]{0.33\textwidth}
        \includegraphics[width=\textwidth]{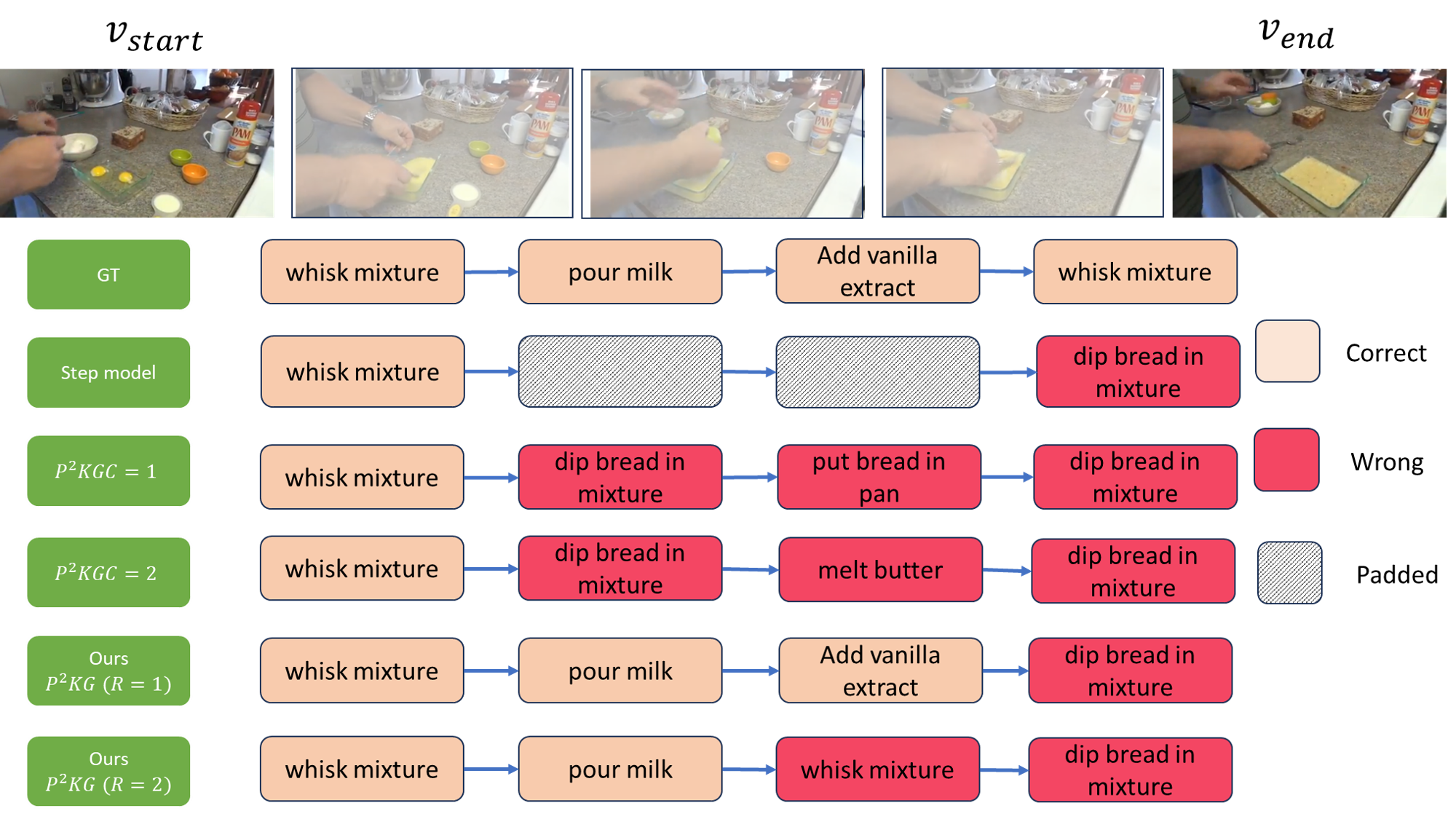}
        \caption{}
        \label{fig:image2}
    \end{subfigure}
    \hfill
    \begin{subfigure}[b]{0.33\textwidth}
        \includegraphics[width=\textwidth]{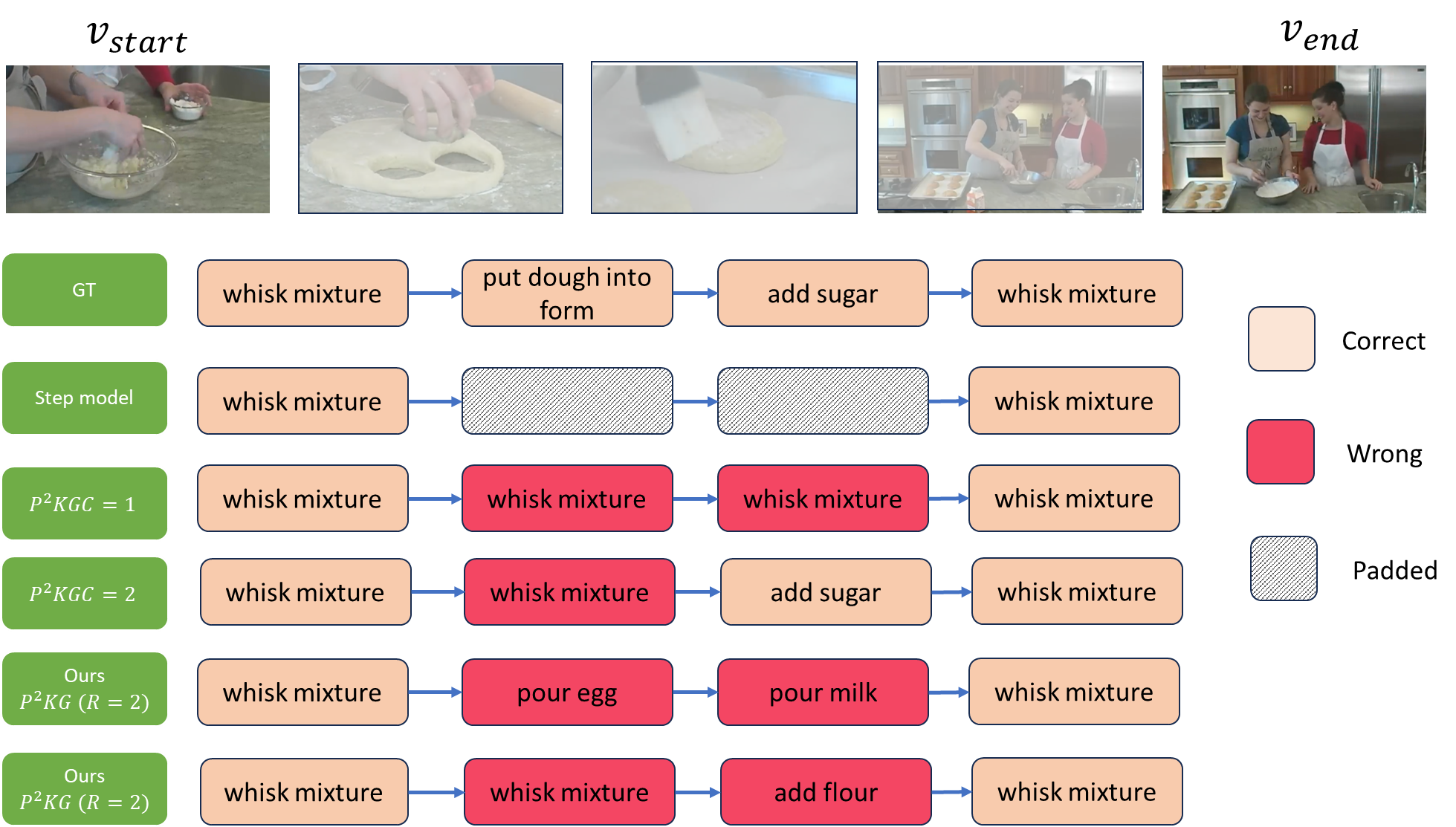}
        \caption{}
        \label{fig:image3}
    \end{subfigure}
    \caption{\textbf{Failure Cases}: (a) when repetitive smaller action sequences exist within the procedure plan, (b) when 
    $a_1$ and $a_T$ have not been accurately predicted by the step model, and (c) when the start and end action steps are the same, and the probabilistic procedure knowledge graph fails to provide valuable plan recommendations because it overly encodes the repetitive actions}
    \label{fig:failure cases}
\end{figure*}
% \fi

\iffalse
\begin{figure*}[htbp]
    \centering
    \begin{subfigure}[b]{0.66\textwidth}
        \includegraphics[width=\textwidth]{fig/failure_1.png}
        \caption{}
        \label{fig:image1}
    \end{subfigure}
    \hfill
    \begin{subfigure}[b]{0.66\textwidth}
        \includegraphics[width=\textwidth]{fig/failure_3.png}
        \caption{}
        \label{fig:image2}
    \end{subfigure}
    \hfill
    \begin{subfigure}[b]{0.66\textwidth}
        \includegraphics[width=\textwidth]{fig/failure_4.png}
        \caption{}
        \label{fig:image3}
    \end{subfigure}
    \caption{\textbf{Failure Cases}: (a) when repetitive smaller action sequences exist within the procedure plan, (b) when 
    % the start and end action steps are the same, and
    $a_1$ and $a_T$ have not been accurately predicted by the step model, and (c) when the start and end action steps are the same, and the probabilistic procedure knowledge graph fails to provide valuable plan recommendations because it overly encodes the repetitive actions}
    \label{fig:failure cases}
\end{figure*}
\fi

\subsection*{B.4 Analysis by the Step Transition Heatmap}

Figure \ref{fig:heatmap} visualises the step transition matrices of the ground-truth training procedure plans, ground-truth testing procedure plans, and procedure plan predictions made by our method with $R$=1 on the test set, for the `Change a Tire' task on the CrossTask dataset. The i-row-j-column depicts the probability of the transition from i-th action step to j-th action step.
% and there are 21 actions illustrated in total. 
Darker color of the step transition heatmap indicates higher probability. 
The steps depicted in the heatmaps in order are: ``brake on'':1, ``get things out'':2, ``start loose'':3, ``jack up'':4, ``unscrew wheel'':5, ``withdraw wheel'':6, ``put wheel'':7, ``screw wheel'':8, ``jack down'':9, ``tight wheel'':10, ``put things back'':11, ``remove funnel'':12, ``lower jack'':13, ``put funnel'': 14, ``wipe off dipstick'': 15, ``close cap'':16, ``insert dipstick'':17, ``pour oil'':18, ``pull out dipstick'':19, ``raise jack'':20, ``remove cap'':21.

Out of the 21 actions illustrated, first 11 actions belong to the `Change a Tire' task while other actions are from several other tasks but highly relevant to the `Change a Tire' task (e.g., `raise jack'':20 and ``lower jack'':13). When comparing Figure \ref{fig:heatmap} (a) and (b), the distributions of action step transitions differ between training and testing.
When comparing Figure \ref{fig:heatmap} (b) and (c), 
our method, which utilizes graph-structured training knowledge, reasonably predicts the step transition matrix on the test set.

\subsection*{B.5 Limitations and Failure Cases}
Figure \ref{fig:failure cases} illustrates three distinct failure case patterns in our model:
(a) failure in prediction when repetitive smaller action sequences exist within the procedure plan, (b) failure in accurately predicting $a_1$ and $a_T$ from the step model, 
and (c) failure in generating valuable probabilistic procedure knowledge graph paths because the graph overly encodes the repetitive actions when the start and end action steps are the same. 
Consistent with prior studies, cases involving repetitive actions or action sequences continue to pose challenges. Future studies could consider addressing these challenging scenarios.

Below, we present analyses regarding how often are errors caused by incorrect $\hat{a}_1$ and $\hat{a}_T$, or the graph. On CrossTask$^\clubsuit$ test data ($T$=4, $R$=2), $49.94\%$ of predicted plans have incorrect $\hat{a}_1$ or $\hat{a}_T$, and $63.23\%$ of error instances feature incorrect $\hat{a}_1$ or $\hat{a}_T$. 
If we use ground-truth $a_1$ and $a_T$ at inference, success rate improves from $21.02$ into $36.58$. When analysing the errors of procedure knowledge graph, in $34\%$ of correctly predicted plans, the top-1 graph plan mismatches the ground-truth plan, and in $87\%$ of error instances, the top-1 graph plan mismatches the ground-truth plan.

The effectiveness of our method is constrained when procedural knowledge is not required. This is because procedural knowledge is structural information about procedures, often considered as \textit{sequential} knowledge about steps in procedures~\cite{zhou2023procedure}, and \ours uses graph \textit{paths} as contextual plan recommendations (and graph paths can provide valuable procedural knowledge), scenarios with short plans restrict the efficacy of \ourseos, which specializes in long planning horizons--the more challenging cases.
% KEPP depends on the precise prediction of $\hat{a}_1$ and $\hat{a}_T$--the smaller the value of $T$, the more pronounced the impact of errors in $\hat{a}_1$ and $\hat{a}_T$ becomes; 
% The smaller the value of $T$, the more KEPP's performance depends on the precise prediction of $\hat{a}_1$ and $\hat{a}_T$ than on the graph or the planning diffusion model.
\ourseos's performance hinges on 
accurate
predictions of $\hat{a}_1$ and $\hat{a}_T$ when $T$ is small, less on the graph for planning;
% or the planning diffusion model 
e.g., with $T$=3, the graph provides only the middle action, and thus procedural knowledge plays a minor role.
% isn't needed.
% the planning diffusion model that leverages graph only predicts the middle action). 
% Conversely, 
% KEPP performs more effectively when $T$ is large, a scenario in which the graph can provide valuable procedural knowledge.
% (\textit{ref.} Line 169-172 in the suppl.).
% a scenario in which the graph can provide \textit{valuable contextual} procedural knowledge (Line 169-172 in the supplement).

% (line 169-172 in the suppl.)

The analyses above imply that performance can be greatly boosted with more precise $\hat{a}_1$ and $\hat{a}_T$,
and having multiple plan recommendations is advantageous, as the top graph path may not always align.
The failure case analyses also highlight the limitations of our approach. Our method depends on the precise prediction of both the initial and final steps by the step model, and it performs more effectively when the probabilistic procedure knowledge graph can offer valuable contextual information.

\subsection*{B.6 Ablations with Flawless Step Model}
The table \ref{tab:GT-table} points out the success rate, mAcc, and mIoU of our plan model for different horizons when the ground truth (GT) $a_1$ and $a_T$ are used for training and testing the model. This provides an accurate depiction of the accuracy of the plans generated when the step (video perception) model generates perfect start and end steps.

\subsection*{B.7 Zero-shot Planning with the P$^2$KG} 
We conduct an experiment by using P$^2$KG for procedure planning as a parameter-free zero-shot approach on CrossTask$^\clubsuit$ ($R$=1 since we directly use top graph plan).
The success rates are $22.58$, $17.74$, $10.92$, and $5.92$ respectively 
for $T$=3,4,5,6; and $56.51$, $32.40$, $19.63$, and $12.16$ 
respectively
if we assume a flawless video model. 

This P$^2$KG-based zero-shot approach could achieve superior performance than current SOTA--for example, SkipPlan's results are $28.85$, $\textit{15.56}$, $\textit{8.55}$, and $\textit{5.12}$, respectively; see Table 1 and Table 2 in the main paper.

\begin{table*}[!htp]
\renewcommand\thetable{S.4} 
\setlength{\tabcolsep}{2.6pt}
\centering
\small
\begin{tabular}{@{}lccc|ccc|ccc|ccc@{}}

\toprule
Model & \multicolumn{3}{c|}{$T$=3} & \multicolumn{3}{c|}{$T$=4} & \multicolumn{3}{c|}{$T$=5} & \multicolumn{3}{c}{$T$=6}
  \\ \cline{2-13}
           &      \( SR^{\uparrow} \) & \( mAcc^{\uparrow} \) & \( mIoU^{\uparrow} \) &      \( SR^{\uparrow} \) & \( mAcc^{\uparrow} \) & \( mIoU^{\uparrow} \) &      \( SR^{\uparrow} \) & \( mAcc^{\uparrow} \) & \( mIoU^{\uparrow} \) &      \( SR^{\uparrow} \) & \( mAcc^{\uparrow} \) & \( mIoU^{\uparrow} \) \\ \midrule
Ours &  33.38 &  60.79 &      63.89 & 21.02 & 56.08 &     64.15 &  12.74 &  51.23 & 63.16&
         9.23& 50.78 & 65.56\\ 
Ours with GT $a_1$, $a_T$  &  57.50 & 84.85 & 80.75 & 37.17 & 77.01 & 78.58 & 20.51 & 67.66 & 73.97 & 15.08 & 65.27 & 74.37\\ 
\bottomrule
\end{tabular}
\caption{Comparison of the performance when ground truth (GT) $a_1$ and $a_T$ are used for training and testing the planning model in our \ours system ($R$=2) on the CrossTask$^\clubsuit$ dataset}
\label{tab:GT-table}
\end{table*}

\subsection*{B.8 Further Discussions}
\noindent \textbf{How can \ours handle novel tasks and scale?}
As mentioned in Sec.4 of the paper, there are trade-offs between using the LLM-generated recommendations and the P$^2$KG recommendations. For instance, the P$^2$KG recommendations are constrained by the training data, limiting their applicability to unseen procedural activities, whereas LLMs generally better generalize to these unseen activities. Nevertheless, it is possible for our proposed method to scale for internet-scale videos given an action list. One could address it by enumerating  all admissible actions and mapping the \ourseos's output actions to the most semantically-similar admissible actions.
% (w/ similarity measure between sentence embeddings).
% KEPP could better generalize and scale when the graph 
% % built at training 
% has a large vocabulary, and thus KEPP would benefit from having internet-scale data to build the graph. 
KEPP generalizes and scales better
% more effectively
with a larger vocabulary in its graph, making internet-scale data beneficial for its development.

\noindent \textbf{Why adapt a diffusion model for step recognition and planning?} 
In the implementation of \ourseos, we adapt PDPP~\cite{wang2023pdpp} which is a diffusion model as the step model and the planning model. We outline our motivations as follows.
% Non-diffusion-based prior models
% (e.g., including Transformer based [46,] and others [9,7,]) 
% often generate step-by-step auto-regressive predictions, leading to error propagation and slow inference. 
First, a diffusion based procedure planning method avoids error propagation seen in non-diffusion-based 
prior 
models~\cite{bi2021procedure,chang2020procedure,sun2022plate}.
% \iffalse
% A diffusion model such as PDPP avoids error propagation 
% % and slow inference
% often associated with step-by-step auto-regressive predictions made by non-diffusion-based prior models such as [9,7,46];
% \fi
Secondly, it can be trained without complex multi-objective training processes~\cite{zhao2022p3iv,wang2023event,li2023skip}.
% \iffalse
% It allows our model to be trained without complicated training process on multiple learning objectives,
% % unique types of loss functions,
% which is the case with [55,49,29];
% \fi
Third, 
% randomness is involved both for training and sampling
% in a diffusion model, 
% helpful to model the uncertainty.
randomness is involved both for training and sampling in a diffusion model, which is helpful to model the uncertain action sequences for planning~\cite{wang2023pdpp}.
% Modeling the uncertainty with a diffusion model 
Furthermore, it is convenient to apply conditional diffusion and projection operations to mitigate uncertainty that might misguide learning.
% misguiding learning;
Finally, the learned weights of the diffusion step and planning models could mutually benefit (e.g., initialize the planning model with the learned step model weights, or fine-tune the step model with the planning model weights).
Future work can consider transfer the learned weights of the diffusion step model towards the diffusion planning model.

\section*{C. Method and Implementation Details}
Our code and trained models are publicly available.

\subsection*{C.1 Diffusion Model Details}

\noindent \textbf{Standard Diffusion Model.}
A standard denoising diffusion probabilistic model~\cite{ho2020denoising} tackles data generation by establishing the data distribution $p\left(x_0\right)$ through a denoising Markov chain over variables $\left\{x_N \ldots x_0\right\}$, starting with $x_N$ as a Gaussian random distribution.

In the forward diffusion phase, Gaussian noise $\epsilon \sim \mathcal{N}(0, \mathbf{I})$ is progressively added to the initial, unaltered data $x_0$, transforming it into a Gaussian random distribution. Each noise addition step is mathematically defined as:
\begin{equation}
    x_n=\sqrt{\bar{\alpha}_n} x_0+\epsilon \sqrt{1-\bar{\alpha}_n}
    \label{eq:x_n}
\end{equation}
\begin{equation}
    q\left(x_n|x_{n-1}\right)=\mathcal{N}\left(x_n ; \sqrt{1-\beta_n} x_{n-1}, \beta_n \mathbf{I}\right)
\end{equation}
where $\bar{\alpha}_n=\prod_{s=1}^n\left(1-\beta_s\right)$ represents the noise magnitude, and $\{\beta_n\in(0,1)\}_{n=1}^N$ denotes the pre-defined ratio of Gaussian noise added in each step.

Conversely, the reverse denoising process transforms Gaussian noise back into a sample. Each denoising step is mathematically defined as:
\begin{equation}
    p_\theta\left(x_{n-1}|x_n\right)=\mathcal{N}\left(x_{n-1} ; \mu_\theta\left(x_n, n\right), \Sigma_\theta\left(x_n, n\right)\right)
\end{equation}
where $\mu_\theta$ is parameterized as a learnable noise prediction model 
$\epsilon_\theta\left(x_n, n\right)$, optimized using a mean squared error (MSE) loss $L=\left\|\epsilon-\epsilon_\theta\left(x_n, n\right)\right\|^2$,
and $\Sigma_\theta$ is calculated using $\left\{\beta_n\right\}_{n=1}^N$.  
During training, the model selects a diffusion step $n \in[1, N]$, calculates $x_n$ via Eq.~\ref{eq:x_n},
then the learnable model
estimates the noise and computes the loss based on the actual noise added at step $n$. After training, the diffusion model generates data akin to $x_0$ by iteratively applying the denoising process, starting from random Gaussian noise.

\noindent \textbf{Conditioned Projected Diffusion Model.} 
As we incorporate conditional information into the data distribution, these conditional guides can be altered during the denoising process. However, modifying these conditions can lead to incorrect guidance for the learning process, rendering conditional guides ineffective. To tackle this issue, the Conditioned Projected Diffusion Model has been introduced to ensure that guided information remains unaffected during denoising. In this approach, Wang et al.~\cite{wang2023pdpp} introduce a condition projection operation within the learning process. Specifically, they enforce that visual observation conditions and additional condition dimensions remain unchanged during both training and inference by assigning them to their initial values. We have adapted their Conditioned Projected Diffusion Model as the architecture for both our step model and planning model in \ourseos.

\subsection*{C.2 Implementation of \ours}
The step model for $a_1$ and $a_T$ predictions and the planning model for procedure plan full-sequence prediction are trained separately.
As mentioned in the main paper, for the step model, intermediate actions are padded in order to accurately predict $a_1$ and $a_T$. 
We use a U-Net based Conditioned Projected Diffusion Model~\cite{wang2023pdpp} $f_\theta(x_n, n)$ as the learnable model architecture for both of the step model and the planning model, and the training loss is:
\begin{equation}
% \small
    \textit{L} =  \sum_{n=1}^{N} ( f_\theta(x_n, n) - x_0 )^2
    \label{eq:train_loss_pjd}
\end{equation}
where $x_0$ denotes the initial, unaltered input and N denotes the total number of diffusion steps. 

In line with the approach taken in \cite{wang2023pdpp}, given that $a_1$ and $a_T$ are the most relevant actions for the provided input visual states, we have modified the training loss by implementing a weighted MSE loss. This involves multiplying the loss by a weight matrix to give more emphasis to the predictions of $a_1$ and $a_T$. For the step model, we have chosen a weight of 10 for both $a_1$ and $a_T$. In the case of the planning model, we have assigned a weight of 5 to $a_1$ and $a_T$ because the condition dimensions of intermediate steps in the probabilistic procedure knowledge graph path has a more significant impact on generating the full action step sequence. The default value in this weight matrix is 1.

The batch size is 256 for all the experiments. Our training regimen incorporates a linear warm-up strategy that is tailored to suit the varying scales of different datasets. For the results of models utilizing the precomputed features provided in CrossTask, we use 200 for the diffusion step and train the progress through a total of 12,000 training steps. The learning rate is gradually escalated to reach 8 × 10$^{-4}$ over the initial 4,000 steps. Subsequent to this phase, we implement a learning rate reduction to 4 × 10$^{-4}$ when reaching the 10,000th step.

When working with the S3D network-extracted features on the CrossTask dataset, the model similarly starts with a diffusion step of 200 but extends the training duration to 24,000 steps. The learning rate here ascends linearly to 5 × 10$^{-4}$ within the first 4,000 steps, followed by a decay factor of 0.5 applied sequentially at the 10,000th, 16,000th, and 22,000th steps.

The NIV dataset, given its smaller size, necessitates a shorter training cycle of 6,500 steps, starting from a diffusion step of 50. We increase the learning rate linearly to 3 × 10$^{-4}$ up until step 4,500, then introduce a decay by 0.5 at step 6,000.

For the COIN dataset, which is significantly larger in scale, the model undergoes an extended training sequence of 160,000 steps with the diffusion set at 200. Here, the learning rate experiences a linear surge to 1 × 10$^{-5}$ within the first 4,000 steps. We implement a decay rate of 0.5 at the 14,000- and 24,000-step marks. After this point, the learning rate is maintained at a constant 2.5 × 10$^{-6}$ for the remainder of the training process.

When obtaining mIoU, for CrossTask dataset we compute IoU on every single action sequence and calculates the average of these IoUs to obtain mean IoU similar to method used in \cite{wang2023pdpp}. For COIN and NIV datasets, we compute mIoU on every mini-batch (batch size = 256) and calculate the average as the result similar to the approach used in \cite{zhao2022p3iv}.

In the scenario where paths from the probabilistic procedure knowledge graph are available between $\hat{a}_1$ and $\hat{a}_T$ but it does not match the desired number of paths ($R$), then repetition of already available paths are considered to meet the desired path requirement.
In scenarios where the combination of $\hat{a}_1$ and $\hat{a}_T$ does not result in the inclusion of a path from the probabilistic procedure knowledge graph, 
the graph paths ($P$) are formulated as follows:
\[ \text{Path Variations ($L$)} = [[[\hat{a}_1]^{T-M} . [\hat{a}_T]^{M}] ,[[\hat{a}_1]^{M} . [\hat{a}_T]^{T-M}]] \]
\[\text{Generated Paths ($P$)} = [L[i\mod \text{len}(L)] \, \text{for} \, i \, \text{in} \, \text{range }(R)]\]
where $T$ is the planning horizon, $M = T//2$, and path variations ($L$) are the two types of possible combinations of $\hat{a}_1$ and $\hat{a}_T$. For example if horizon (T) equals 5, then the path variations (L) possible are [[$\hat{a}_1$, $\hat{a}_1$, $\hat{a}_1$, $\hat{a}_T$, $\hat{a}_T$], [$\hat{a}_1$, $\hat{a}_1$, $\hat{a}_T$, $\hat{a}_T$, $\hat{a}_T$]]. So if one path ($R=1$) is required as the condition for the planning model, generated path is [$\hat{a}_1$, $\hat{a}_1$, $\hat{a}_1$, $\hat{a}_T$, $\hat{a}_T$], and if 2 paths are required ($R=2$) as the condition for the planning model, generated paths are [$\hat{a}_1$, $\hat{a}_1$, $\hat{a}_1$, $\hat{a}_T$, $\hat{a}_T$], and [$\hat{a}_1$, $\hat{a}_1$, $\hat{a}_T$, $\hat{a}_T$, $\hat{a}_T$], and if 3 paths are required ($R=3$) as the condition for the planning model, generated paths are [$\hat{a}_1$, $\hat{a}_1$, $\hat{a}_1$, $\hat{a}_T$, $\hat{a}_T$], [$\hat{a}_1$, $\hat{a}_1$, $\hat{a}_T$, $\hat{a}_T$, $\hat{a}_T$], and [$\hat{a}_1$, $\hat{a}_1$, $\hat{a}_1$, $\hat{a}_T$, $\hat{a}_T$] likewise. 
The generated paths ($P$) are then aggregated through linear weighting into a single path to be given to the planning model as mentioned in Section
% ~\ref{subsubsec:param_sensitivity}
\red{A.2}.

\subsection*{C.3 Implementation of Ablations}

\noindent \textbf{Ablation on the probabilistic procedure knowledge
graph.} When the probabilistic procedure knowledge graph is not included as a condition, the model simplifies only to the planning model, ignoring the step model for generating $\hat{a}_1$ and $\hat{a}_T$. In this scenario, the planning model generates the action sequences using a diffusion process conditioned solely on the start and end visual observations.

\noindent \textbf{Plan recommendations provided by the probabilistic procedure knowledge graph and LLM.} To enhance the action prediction, we could incorporate LLM generated action sequence as a condition to our planning diffusion model as shown in Table 6 in the main paper. To generate the LLM action sequence we use the `llama-2-13b-chat' or the `llama-2-70b-chat' model and query the model with the prompt:

% (‘llama-2-13b-chat’)
\bigskip 

``\texttt{In the multi-step task that I am going to perform, I know \{total\_steps\} steps are required, the first step is \{start\_action\}, and the last step is \{end\_action\} and I need to go from \{start\_action\} to \{end\_action\}. Can you tell me the \{total\_steps\} steps in the form ”Step 1, ..., Step \{total\_steps\}” so that I can follow in order to complete this \{total\_steps\}-step task? Note that repetitive steps are possible}''. 

\bigskip 

In the above text prompt, `total\_steps' indicates the planning horizon, while `start\_action' denotes the step model-predicted start action $\hat{a}_1$ and `end\_action' denotes the step model-predicted end action $\hat{a}_T$. For a single [$\hat{a}_1, \hat{a}_T$] pair occurring in the training or testing phase, we query the LLM three times and obtain the optimal action for each step in the sequence by finding the action with the highest occurrence. Then, the optimal sequence is queried back into the LLM for post-processing to check the realistic nature of the generated action sequence.
The text prompt used for post-processing verification is as follows:

\bigskip 

``\texttt{Is the following action sequence possible? Note that repetitive steps are possible. Please answer only ”Yes” or ”No”: \{steps\}}''.

\bigskip 

Here, `\{steps\}' denotes the the generated steps from the first prompt. 
We then map the action steps generated from the LLM to the step vocabulary space of the given dataset. Specifically, the generated action steps from the LLM, as well as all of the possible action steps from the dataset, are embedded into a vector space utilizing the `bert-base-uncased' model. For each action in the LLM-generated sequence, its semantically closest action step from the dataset is found in the vector space using the K-nearest neighbor (K=1) approach.
The resulting generated action sequence is given as an additional input condition to the planning model similar to the P$^2$KG condition.

In the situation where both the P$^2$KG and LLM conditions are used for the planning model, the conditional visual states, LLM recommendation and the procedure plan recommendation from the P$^2$KG are concatenated with the actions along the action feature dimension, forming a multi-dimensional array:

\begin{equation}
    \left[\begin{array}{ccccc}
\cellcolor{gray!25}v_s & \cellcolor{gray!25}0 & \cellcolor{gray!25}\ldots & \cellcolor{gray!25}0 & \cellcolor{gray!25}v_g\\
\cellcolor{gray!25}\tilde{a}_1 & \cellcolor{gray!25}\tilde{a}_2 & \cellcolor{gray!25}\ldots & \cellcolor{gray!25}\tilde{a}_{T-1} & \cellcolor{gray!25}\tilde{a}_T\\
\cellcolor{gray!25}a^*_1 & \cellcolor{gray!25}a^*_2 & \cellcolor{gray!25}\ldots & \cellcolor{gray!25}a^*_{T-1} & \cellcolor{gray!25}a^*_T\\
a_1 & a_2 & \ldots & a_{T-1} & a_T
\end{array}\right]
% \vspace{-20pt}
\end{equation}
where $\tilde{a}_i$ denotes the action from the P$^2$KG at $i^{th}$ step and $a^*_i$ denotes the action generated from LLM at $i^{th}$ step.

\noindent  \textbf{Probabilistic procedure knowledge graph (P$^2$KG) and Frequency-based procedure knowledge graph (PKG).} The frequency-based procedure knowledge graph (PKG) performs min-max normailization for the entire graph, and the optimal path between two nodes is the maximum weighted path. The path weight is calculated by summing all the individual edge weights between the two nodes. The probabilistic procedure knowledge graph (P$^2$KG) has undergone output edge normalization, where the summation of probabilities of all the output paths from a node is 1. For example, if a node has four output edges (self-loop edges are also considered as output edges), and the weights along the edges are $w_1, w_2, w_3, w_4$, then the probability of each edge is:
\vspace{1mm}
\fontsize{8}{12}\selectfont
$
\frac{w_1}{w_1 + w_2 + w_3 + w_4} , \frac{w_2}{w_1 + w_2 + w_3 + w_4} , \frac{w_3}{w_1 + w_2 + w_3 + w_4} , \frac{w_4}{w_1 + w_2 + w_3 + w_4}
$ 
\normalsize 
respectively. The process of obtaining the highest probable path between two nodes is explained in Section 3.2.3 of the main paper.

\subsection*{C.4 Datasets and Evaluation Metrics}
In our evaluation, we employed datasets from three sources: CrossTask \cite{zhukov2019cross}, COIN \cite{tang2019coin}, and the Narrated Instructional Videos (NIV) \cite{alayrac2016unsupervised}. The CrossTask dataset comprises 2,750 video clips, each representing one of 18 distinct tasks, and features an average of $7.6\pm4.4$ (mean$\pm$std) actions per clip. The COIN dataset is more extensive, including 11,827 videos across 180 tasks, with an average of $3.9\pm2.4$ actions per video. Lastly, the NIV dataset, though smaller in scale, includes 150 videos that capture 5 everyday tasks, with a higher density of actions, averaging $8.8\pm2.8$ actions per video. We randomly select 70\% data for training and 30\% for testing as previous work \cite{wang2023pdpp, bi2021procedure, zhao2022p3iv}.

Our study employs mean intersection over union (mIoU), mean accuracy (mAcc),  and success rate (SR) as our evaluation metrics.
mIoU evaluates the overlap between
the predicted actions with the ground-truth actions using IoU by regarding them as two action sets.
mAcc is quantified as the average of the precise correspondences between the predicted actions and their respective ground truth counterparts at corresponding time steps. 
Thus, mAcc counts the order of actions, and is stricter than mIoU. \textbf{SR is the most stringent metric}, considering a sequence to be successful solely if it exhibits a perfect match with the ground truth action sequence throughout.

\subsection*{C.5 Baselines}

In the current study, we benchmark our proposed approach against several established state-of-the-art methods to conduct a comprehensive evaluation. Below, we describe
the methods that serve as our baseline comparisons.

\textbf{Random Strategy}: This naive approach involves the stochastic selection of actions from the set of possibilities within the dataset to generate procedure plans.

\textbf{Retrieval-based Technique}: Upon receiving the visual observational inputs, this method employs a nearest-neighbour algorithm, leveraging the least visual feature space distance within the training corpus to extract a corresponding sequence of actions as the procedural blueprint.

 \textbf{WLTDO} \cite{ehsani2018let} [`CVPR 2018']: Implemented via a recurrent neural network (RNN), this method sequentially forecasts the necessary actions based on the supplied pair of observations.

\textbf{UAAA} \cite{abu2019uncertainty} [`ICCV workshop 2019']: Standing for a two-phased strategy, UAAA integrates an RNN with a Hidden Markov Model (HMM) to autoregressively estimate the sequence of actions.

 \textbf{UPN} \cite{srinivas2018universal} [`ICML 2018']: This path-planning algorithm is tailored for the physical realm and acquires a plannable representational form for making informed predictions. 

 \textbf{DDN} \cite{chang2020procedure} [`ECCV 2020']: The DDN framework operates as a dual-branch autoregressive model, honing in on an abstracted representation of action sequences to anticipate the transitions between states and actions in the feature space.

 \textbf{PlaTe} \cite{sun2022plate} [`RA-L 2022']: An advancement of DDN, the PlaTe model incorporates transformer modules within its dual-branch architecture to facilitate prediction. 

\textbf{Ext-GAIL} \cite{bi2021procedure} [`ICCV 2021']: Ext-GAIL addresses procedural planning through reinforcement learning and decomposes the problem into a dual-stage challenge. The initial stage in Ext-GAIL is designed to yield time-invariant contextual
% extensive horizon
information for the subsequent planning phase.

 \textbf{P$^3$IV} \cite{zhao2022p3iv} [`CVPR 2022']: A transformative single-branch model,  P$^3$IV enriches its structure with a mutable memory repository and an additional generative adversarial network.  It is capable of concurrently predicting the entire action sequence during the inference stage.

\textbf{PDPP} \cite{wang2023pdpp} [`CVPR 2023']: PDPP conceptualizes the task of procedure planning in instructional videos as analogous to fitting a probabilistic distribution. 
Echoing the P$^3$IV, PDPP employs a holistic approach during the inference stage by forecasting the entire sequence of action steps concurrently.

\textbf{SkipPlan} \cite{li2023skip} [`ICCV 2023']: Consider procedure planning as a mathematical chain problem. This model decomposes relatively long chains into multiple short sub-chains by bypassing unreliable intermediate actions. This transfers long and complex sequence functions into short but reliable ones.

\textbf{E3P} 
% / \textbf{EGPP}
\cite{wang2023event} [`ICCV 2023']: This uses procedural task (referred as `event' in \cite{wang2023event}) information for procedure planning by encoding task information into the sequential modeling process. E3P infers the task information from the observed states and then plans out actions based on both the states and the predicted task/event.

\end{document}